\documentclass[conference]{IEEEtran}
\IEEEoverridecommandlockouts
\usepackage{cite}
\usepackage[english]{babel}

\usepackage[T1]{fontenc}
\usepackage{multirow}
\usepackage[table,xcdraw]{xcolor}
\usepackage{amsmath,amssymb,amsfonts}
\usepackage{multicol}
\usepackage{graphicx}
\graphicspath{{images/}}
\usepackage{color}
\usepackage{textcomp}
\usepackage{soul}
\usepackage{comment}
\usepackage{caption}
\usepackage{subcaption}

\sethlcolor{yellow}
\usepackage[colorinlistoftodos,prependcaption,textsize=normalsize]{todonotes}
\usepackage{xargs}
\usepackage[colorlinks=true, allcolors=blue]{hyperref}
\def\BibTeX{{\rm B\kern-.05em{\sc i\kern-.025em b}\kern-.08em
        T\kern-.1667em\lower.7ex\hbox{E}\kern-.125emX}}
\DeclareUnicodeCharacter{2212}{-}
\begin{document}

\title{Deep Learning based Systems for Crater Detection: A Review}

\author{\IEEEauthorblockN{Atal Tewari}
\IEEEauthorblockA{Electrical Engineering \\
IIT Gandhinagar, India \\
atal.tewari@iitgn.ac.in}
\and
\IEEEauthorblockN{K Prateek}
\IEEEauthorblockA{Electrical Engineering \\
IIT Gandhinagar, India \\
k\_prateek@alumni.iitgn.ac.in}
\and
\IEEEauthorblockN{Amrita Singh}
\IEEEauthorblockA{Electrical Engineering \\
IIT Gandhinagar, India \\
amiritas@iitgn.ac.in}
\and
\IEEEauthorblockN{Nitin Khanna}
\IEEEauthorblockA{Electrical Engineering \\
IIT Bhilai, India \\
nitin@iitbhilai.ac.in}

}
\maketitle
\begin{abstract}
Craters are one of the most prominent features on
planetary surfaces, used in applications such as age estimation, hazard detection, and spacecraft navigation. 
Crater detection is a challenging problem due to various aspects, including complex crater characteristics such as varying sizes and shapes, data resolution, and planetary data types.
Similar to other computer vision tasks, deep learning-based approaches have significantly impacted research on crater detection in recent years.
This survey aims to assist researchers in this field by examining the development of deep learning-based crater detection algorithms (CDAs). The review includes over $140$ research works covering diverse crater detection approaches, including planetary data, craters database, and evaluation metrics. 
To be specific, we discuss the challenges in crater detection due to the complex properties of the craters and survey the DL-based CDAs by categorizing them into three parts: (a) semantic segmentation-based, (b) object detection-based, and (c) classification-based. 
Additionally, we have conducted training and testing of all the semantic segmentation-based CDAs on a common dataset to evaluate the effectiveness of each architecture for crater detection and its potential applications.
Finally, we have provided recommendations for potential future works.
\end{abstract}
\maketitle

\section{Introduction}

Craters are one of the prominent topographic features on most planetary surfaces.
Several missions have been carried out over the last few decades to explore the planetary surfaces, which helps to understand the physical properties of planetary surfaces and how impact rates vary over time~\cite{goswami2011chandrayaan,zurek2007overview,xiao2021chang,chicarro2004mars,kato2010kaguya}.
The spatial and size-frequency distribution of craters is critical for understanding the impactor population as well as collisional and evolutionary events in the solar system, which helps to derive the impact flux of the solar system~\cite{strom2005origin,fassett2016analysis,toyokawa2022kilometer,lin2022lunar}. 
Most notably, impact craters are studied to understand impactor energy, angle, mechanical properties of a target's regolith, projectile type, size, and other factors that influence the morphologies of these craters, such as material strength and gravity~\cite{gault1974impact,kieffer1980role,liu2019machine}.
Additionally, it is also used in applications of space probes like landform selection and spacecraft navigation~\cite{yu2014new}.

With the continuous advancement of technology, high-resolution data is now available, and identifying a wide size range of craters is possible. A crater can be detected either manually or automatically. In the manual approach, domain experts visually inspect data to annotate craters. Some previous works, such as Robbins et al.~\cite{robbins2019new} and Head et al.~\cite{head2010global}, use a manual approach to mark the craters. However, manual marking is cumbersome, time-consuming, and prone to error. Robbins et al.~\cite{robbins2014variability} stated that a $\sim$ $45\%$ discrepancy exists in marking craters among experts. Therefore, most studies followed the automatic approach for crater detection to reduce human time and biases in manual marking.

Many researchers have widely considered the automatic crater detection approach to develop an efficient and accurate detection algorithm. 
The developed automatic crater detection algorithms (CDAs) can be divided into two types:  traditional and deep learning (DL) based methods. 
Traditional CDAs (e.g.~\cite{kim2005automated,martins2009crater,bandeira2012detection,stepinski2009machine,yamamoto2015rotational,zhou2018automatic}) typically first extracted the handcrafted features such as edge, contour, and depression and then utilized these features to detect the craters.
For example, Kim et al.~\cite{kim2005automated} extracted the edge features and then used the template matching to find the final craters. However, traditional crater detection methods are not generalizable to larger surface areas and a wide diameter range~\cite{silburt2019lunar,li2017recognizing}.

Recently deep learning has been successful in various computer vision tasks such as object detection~\cite{ren2015faster,wang2021end,carion2020end,dai2021dynamic} and semantic segmentation~\cite{ronneberger2015u,sun2019deep,fu2020scene,cheng2020cascadepsp}.
In contrast to traditional machine-learning models that rely on handcrafted features, deep learning models have the ability to learn and identify features on their own. 
As a result, in the last few years, most crater detection works have followed deep learning-based approaches.

In this work, we comprehensively review the important aspects of existing deep learning-based crater detection algorithms (CDAs)  and mention some recent advancements in the field. 
Based on computer vision tasks, we categorized deep learning-based CDAs based on their approach into three types: semantic segmentation-based, object detection-based, and classification-based (Figure~\ref{fig:Survey_work_categorization}).
In semantic segmentation-based CDAs, semantic segmentation-based deep neural networks (DNNs) are used to categorize each pixel in the image as crater or non-crater. 
Then, a non-deep learning (DL) method, such as template matching is used to extract the location and size of the craters.
Object detection-based CDAs directly provide the craters' location and size information by utilizing the object detection frameworks such as Faster R-CNN~\cite{ren2015faster}. 
In classification-based CDAs, features are extracted using traditional methods (i.e., non-deep learning techniques) and subsequently classified into the crater and non-crater categories using DNNs.
We discuss the significance of various CDAs that can be further utilized in research. 
Therefore, this study tries to understand and analyze a more holistic approach to provide a stronger foundation for developing crater detection algorithms (CDAs). 
Also, describe the challenge due to variation of crater features, dataset description for crater detection, and finally, explain some key points that need to be cognizant for future crater detection work.

\section{Scope and Aim}
\label{sec:Scope and Aim}

Based on the most recent development trend of deep learning (DL) based CDA, the performance of the algorithms is continuously improving to meet the demand of various scientific applications.
This paper aims to examine the evolution of deep learning-based automatic crater detection algorithms (CDAs).
Also, we re-implement the semantic segmentation-based CDAs on a generated benchmark dataset to understand their effectiveness in terms of accuracy and speed.
The major contributions of our research are as follows:
\begin{itemize}
    \item Based on the computer vision task, we categorize the deep learning-based crater detection method and summarize its main characteristics, research approaches, and limitations.  

    \item 

    Different existing works utilize data and catalogs based on their accessibility and convenience. All existing semantic segmentation-based crater detection works are trained, evaluated, and analyzed on common data, catalog, and diameter range to understand the effectiveness of the deep learning framework. It will help scientists determine which architecture to employ for their application.

    \item We have presented a detailed description of the CDAs dataset and catalog to assist researchers in determining which ones best suit their requirements and objectives. Specifically, we have provided information on each dataset and catalog used in CDAs, such as resolution, coverage area, and detected crater diameter range.
    
    \item We have provided the challenges and problems encountered in the crater detection field, such as the complexity of crater features and the lack of labeled data. Further, a set of promising future works and recommendations are provided that could determine the direction of forthcoming developments and advancements of CDAs.
\end{itemize}

A number of papers on crater detection have been published, including both manual and automatic methods. 
This work focuses on papers that use deep learning-based crater detection approaches and have been published in peer-reviewed journals before February 2022.
In order to find pertinent publications in google scholar, we combined several keywords, including "crater counting", "crater detection algorithm", "convolutional neural network", and "deep learning." 
We sincerely apologize to the authors whose works were based on crater detection but are not included in this review.

The remaining paper is organized as follows. Section~\ref{sec:Comparison With Previous Review} provides a comparison with previous review work. Section~\ref{sec:Challenges for crater detection} discusses the challenges associated with crater detection. Section~\ref{sec:Categorization of Deep learning based method for crater detection} is divided into three major sub-sections, each focusing on a different type of deep learning-based CDAs. The first Section~\ref{sec:Semantic Segmentation Based Method} discusses the semantic segmentation-based crater detection methods. It is followed by a comparative analysis of frameworks, dataset preparation, implementation details, results, and discussions. 
In the second Section~\ref{sec:Object Detection Based Method}, a brief background on object detection-based CDAs and terms frequently used in object detection are provided. This section contains definitions and a review of the proposed methods. 
Finally, Section~\ref{sec:Classification Based Method} describes the classification-based crater detection method. 
Section~\ref{sec:Description of Benchmark Dataset} describes the various benchmark datasets. 
Section~\ref{sec:Future Direction} examines the future trends and provides further research directions, while Section~\ref{sec:Conclusion} concludes the paper.

\section{Comparison With Previous Review}
\label{sec:Comparison With Previous Review}

Currently, there is only one survey paper for the deep learning-based crater detection methods, published in 2019 by DeLatte et al.~\cite{delatte2019automated}. Our work has numerous enhancements in contrast with their paper. 
First, we also included papers published after $2019$, and more than $140$ papers are cited in this review.
Second, to the best of our knowledge, this is the first literature review paper that benchmark semantic segmentation-based crater detection frameworks on a common dataset. 
The result obtained by comparing the effectiveness of the different automatic CDAs on a common dataset aids in gaining insight into each algorithm's key features, comprehending how they differ from one another, and also help researchers in selecting the best automatic CDA architecture suited to their needs and application. 
Third, this review paper clarifies the similarity and differences between different deep learning frameworks used for crater detection. Furthermore, we have provided a tabular representation of the key features and characteristics of different deep learning-based crater detection methods given in Table \ref{tab:cda_key_info} and a detailed description of the data and catalog used in previous work given in Table \ref{tab:catalog} and Table~\ref{tab:data}, respectively.

\section{Challenges for crater detection}
\label{sec:Challenges for crater detection}

\begin{figure*}[!ht]
     \centering
     \begin{subfigure}[b]{0.3\textwidth}
         \centering
         \includegraphics[width=\textwidth]{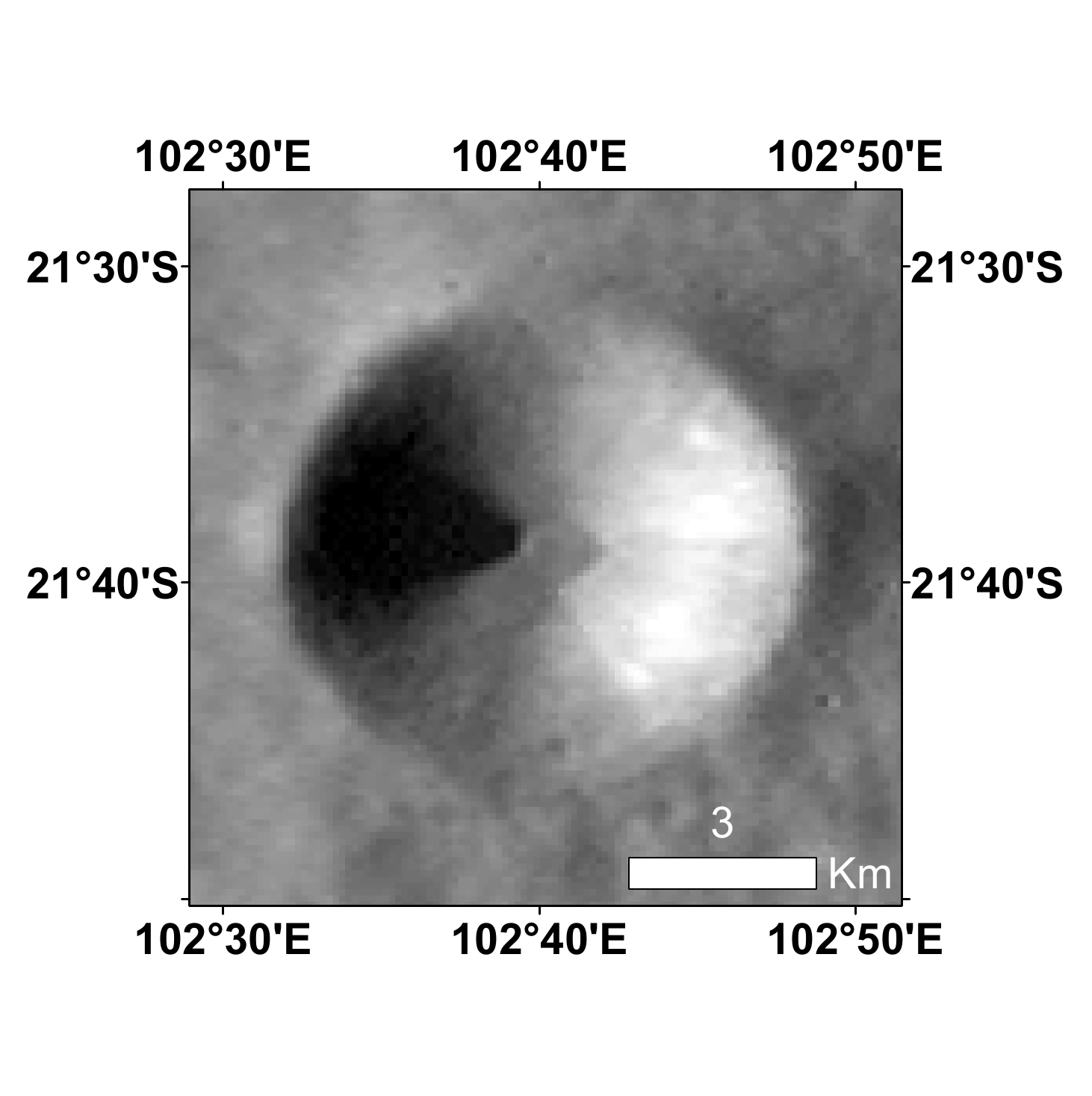}
         \caption{Circular}
     \end{subfigure}
     \begin{subfigure}[b]{0.3\textwidth}
         \centering
         \includegraphics[width=\textwidth]{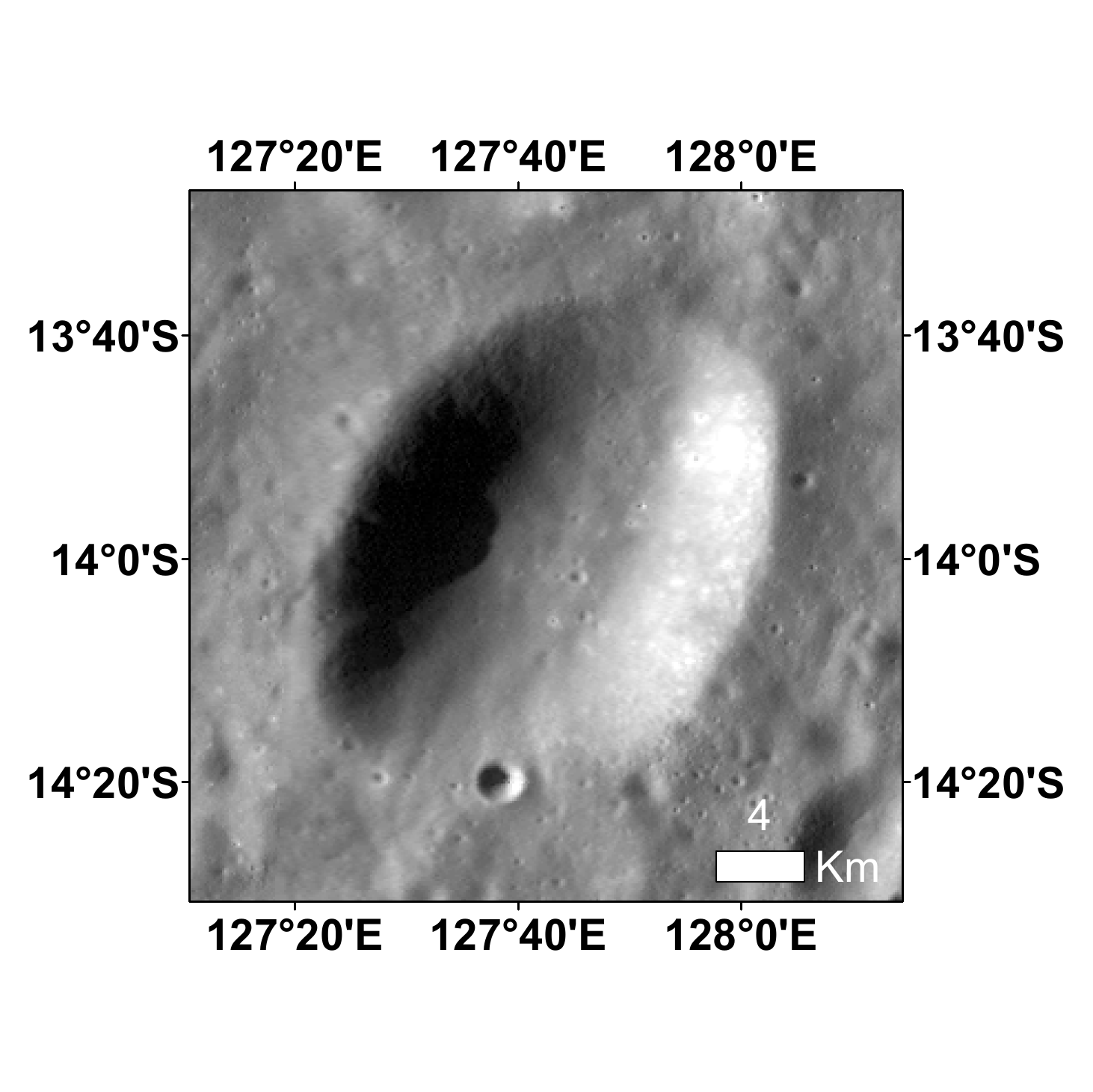}
         \caption{Elliptical}
     \end{subfigure}
     \begin{subfigure}[b]{0.3\textwidth}
         \centering
         \includegraphics[width=\textwidth]{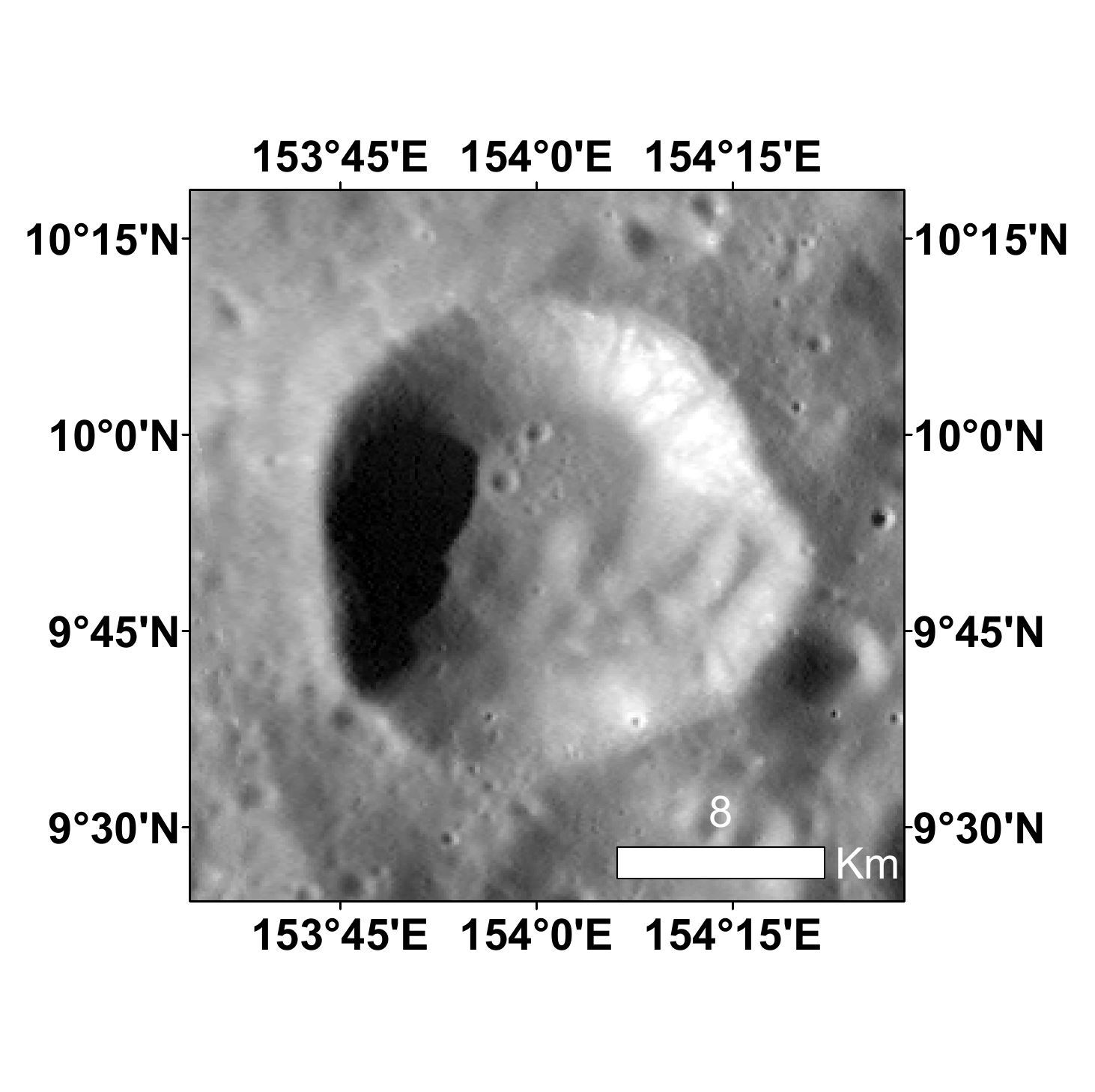}
         \caption{Irregular}
     \end{subfigure}
     \begin{subfigure}[b]{0.3\textwidth}
         \centering
         \includegraphics[width=\textwidth]{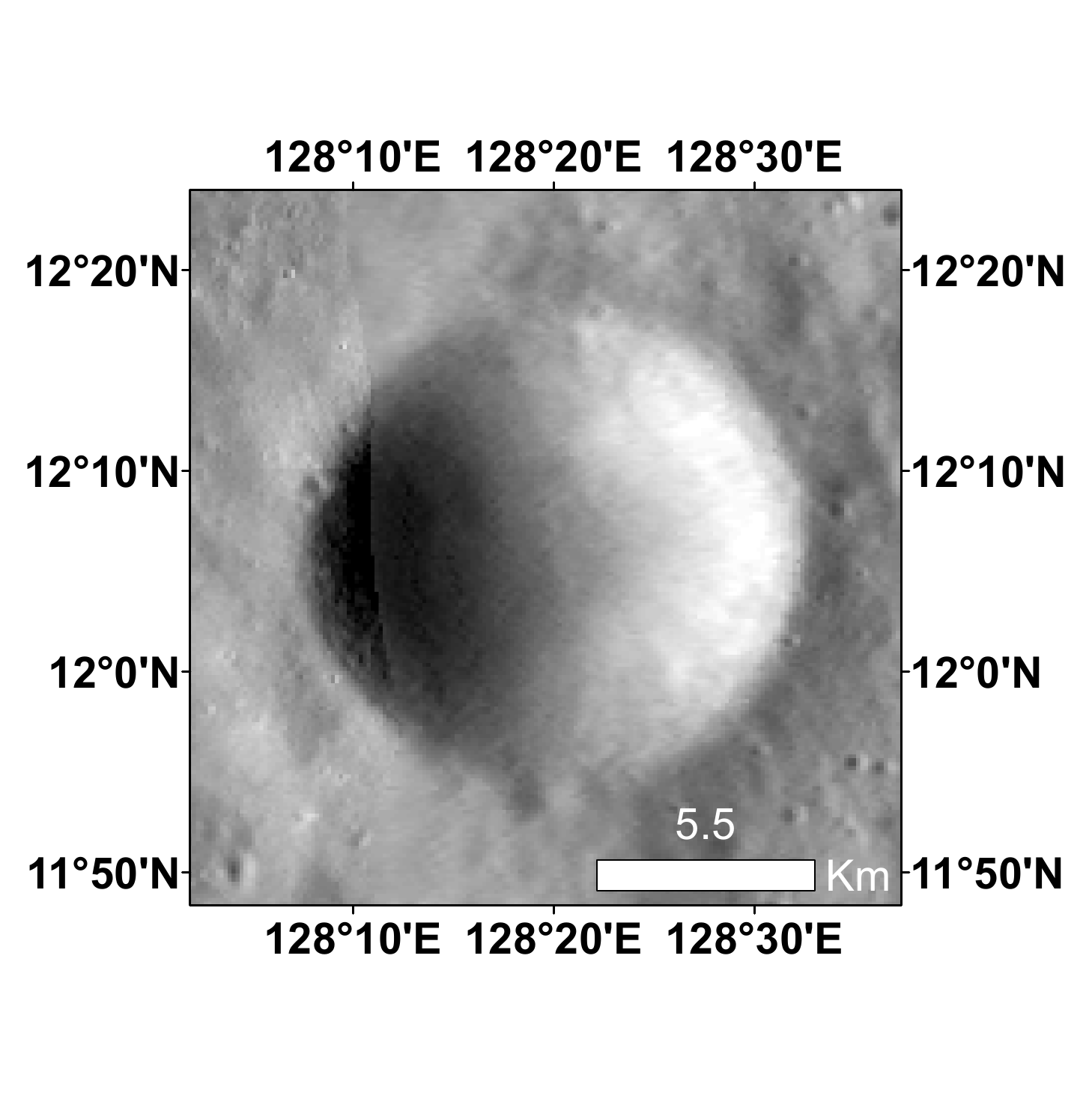}
         \caption{Round}
     \end{subfigure}
     \begin{subfigure}[b]{0.3\textwidth}
         \centering
         \includegraphics[width=\textwidth]{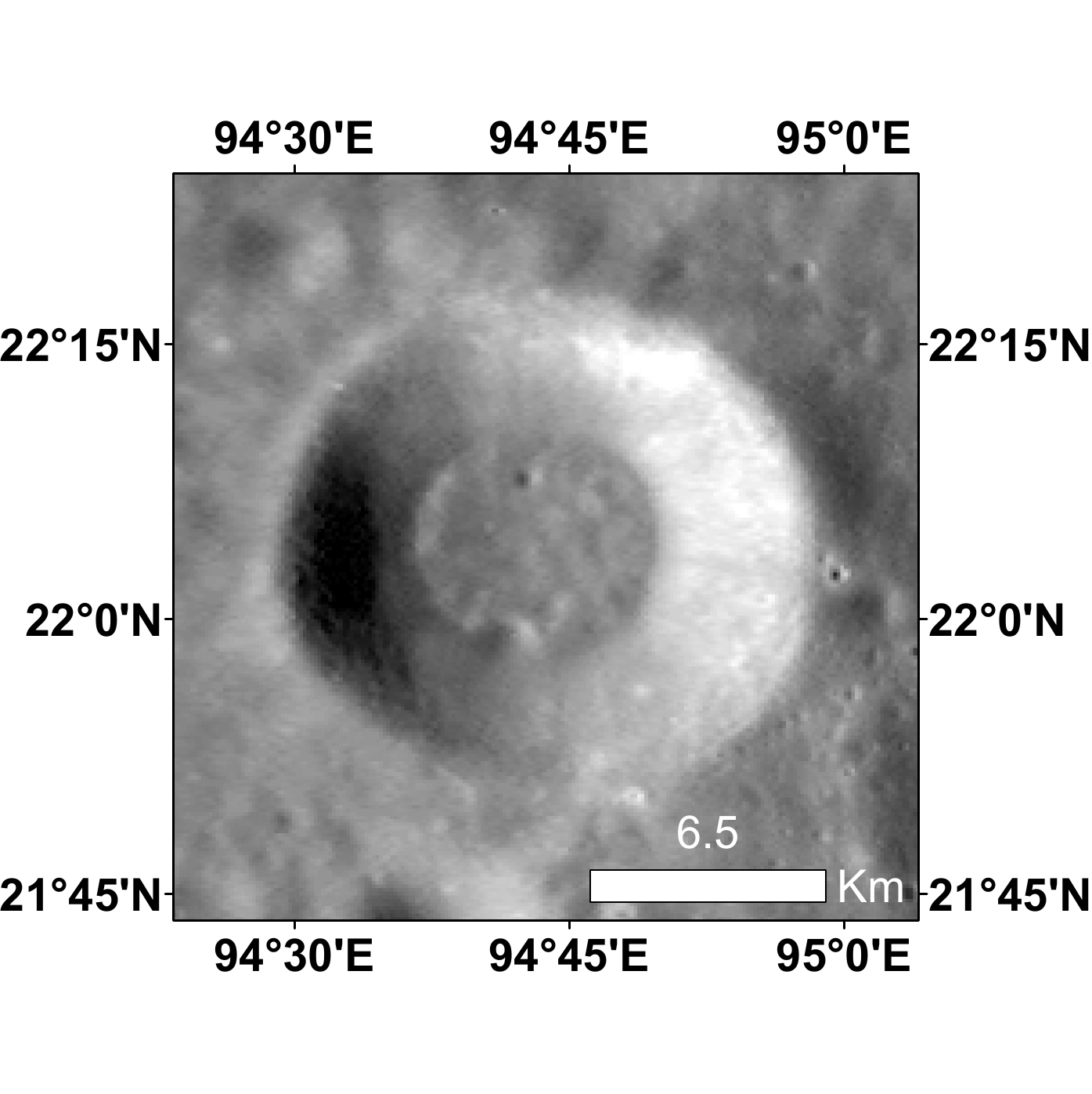}
         \caption{Flat}
     \end{subfigure}
     \begin{subfigure}[b]{0.3\textwidth}
         \centering
         \includegraphics[width=\textwidth]{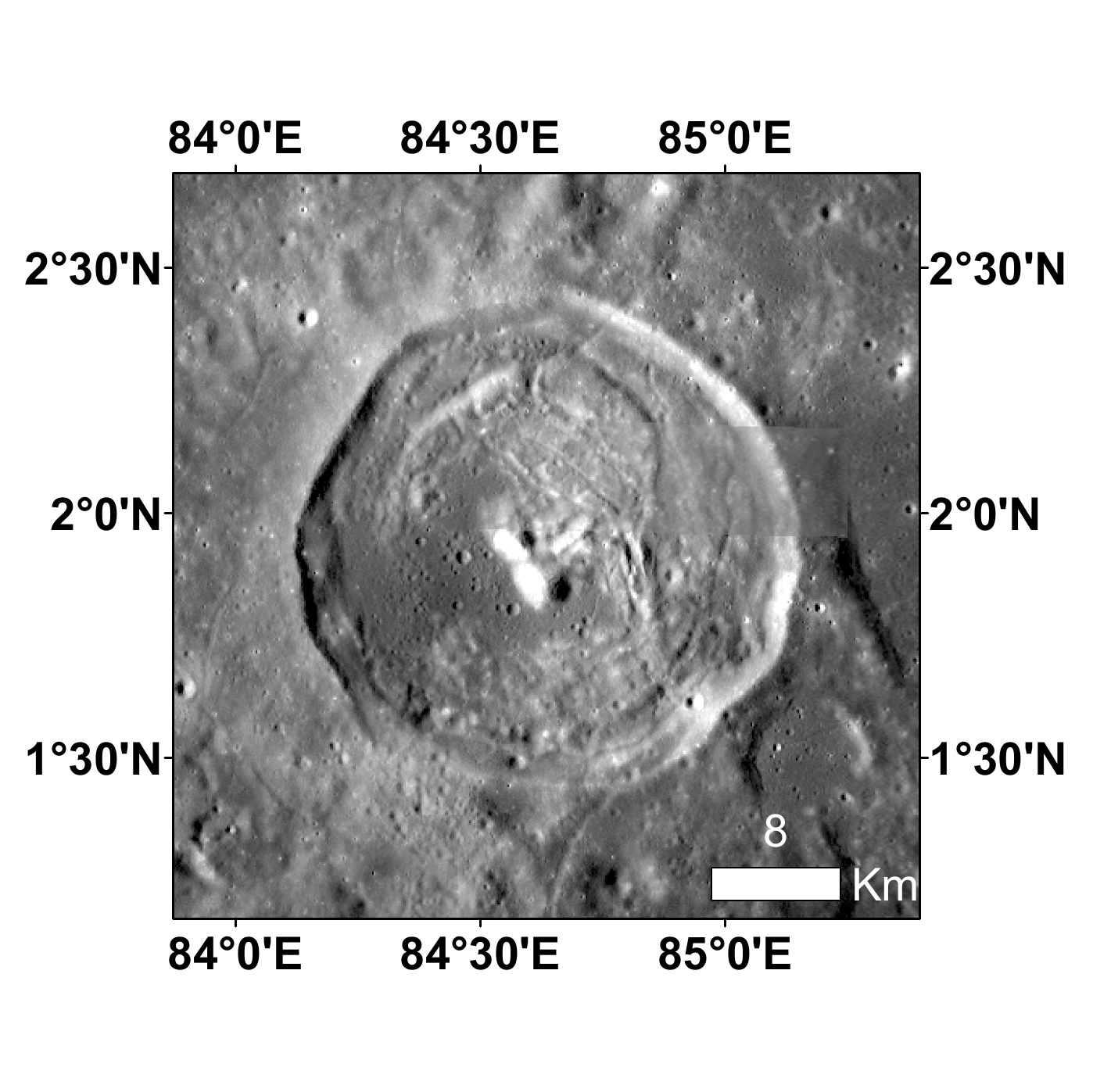}
         \caption{Central Pit}
     \end{subfigure}
     \begin{subfigure}[b]{0.3\textwidth}
         \centering
         \includegraphics[width=\textwidth]{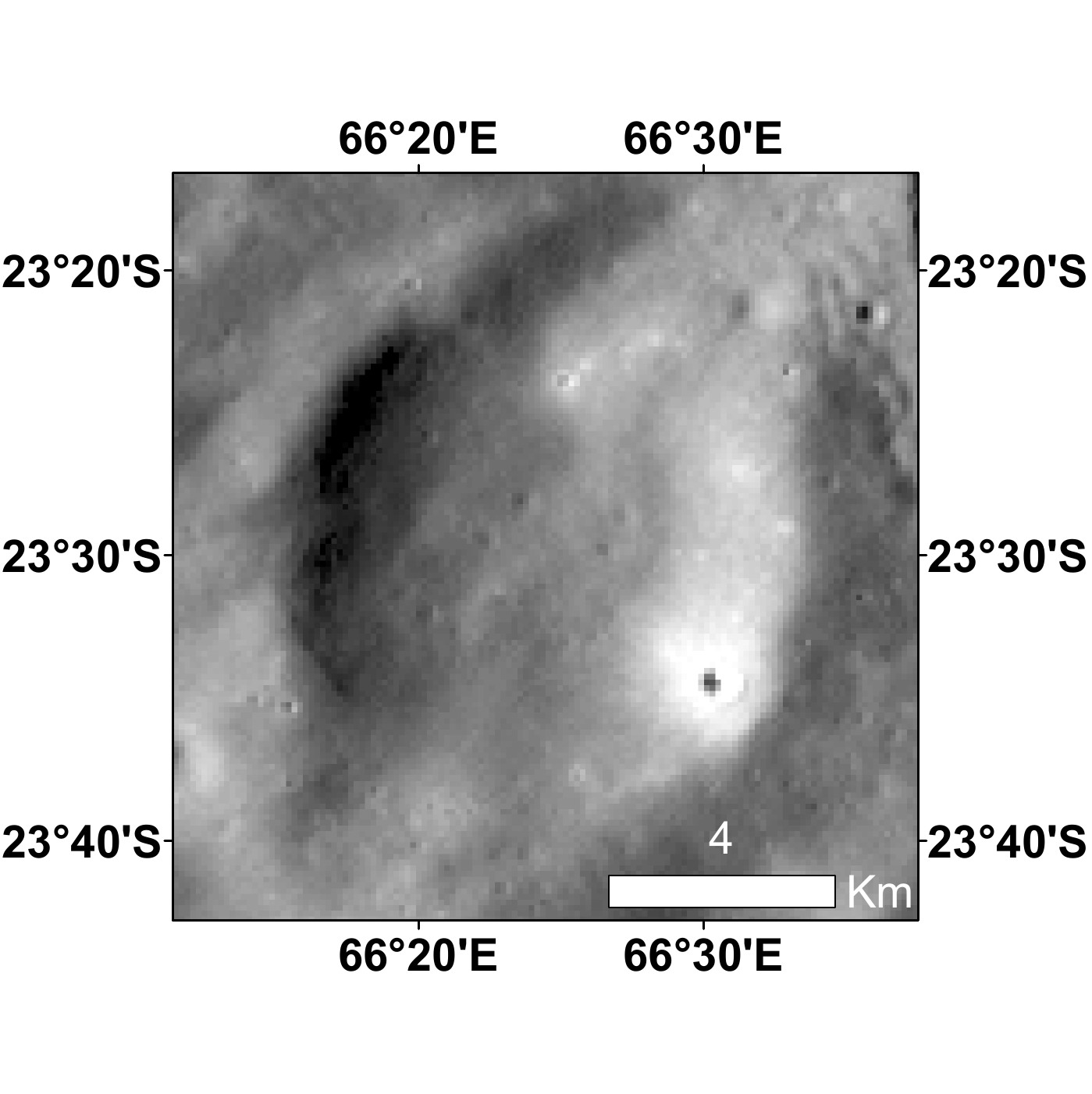}
         \caption{Degraded}
     \end{subfigure}
     \begin{subfigure}[b]{0.3\textwidth}
         \centering
         \includegraphics[width=\textwidth]{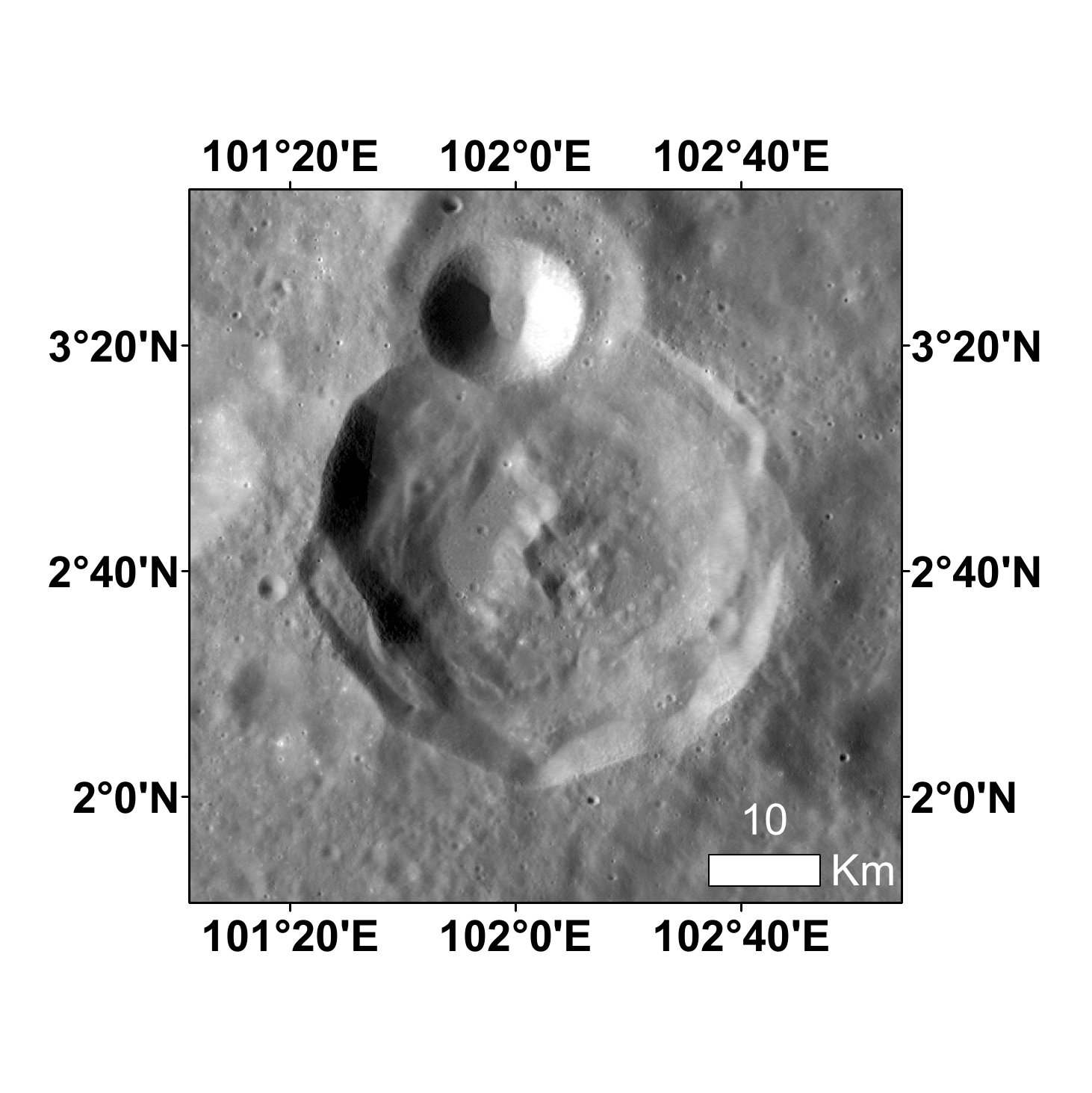}
         \caption{Overlapping}
     \end{subfigure}
     \begin{subfigure}[b]{0.3\textwidth}
         \centering
         \includegraphics[width=\textwidth]{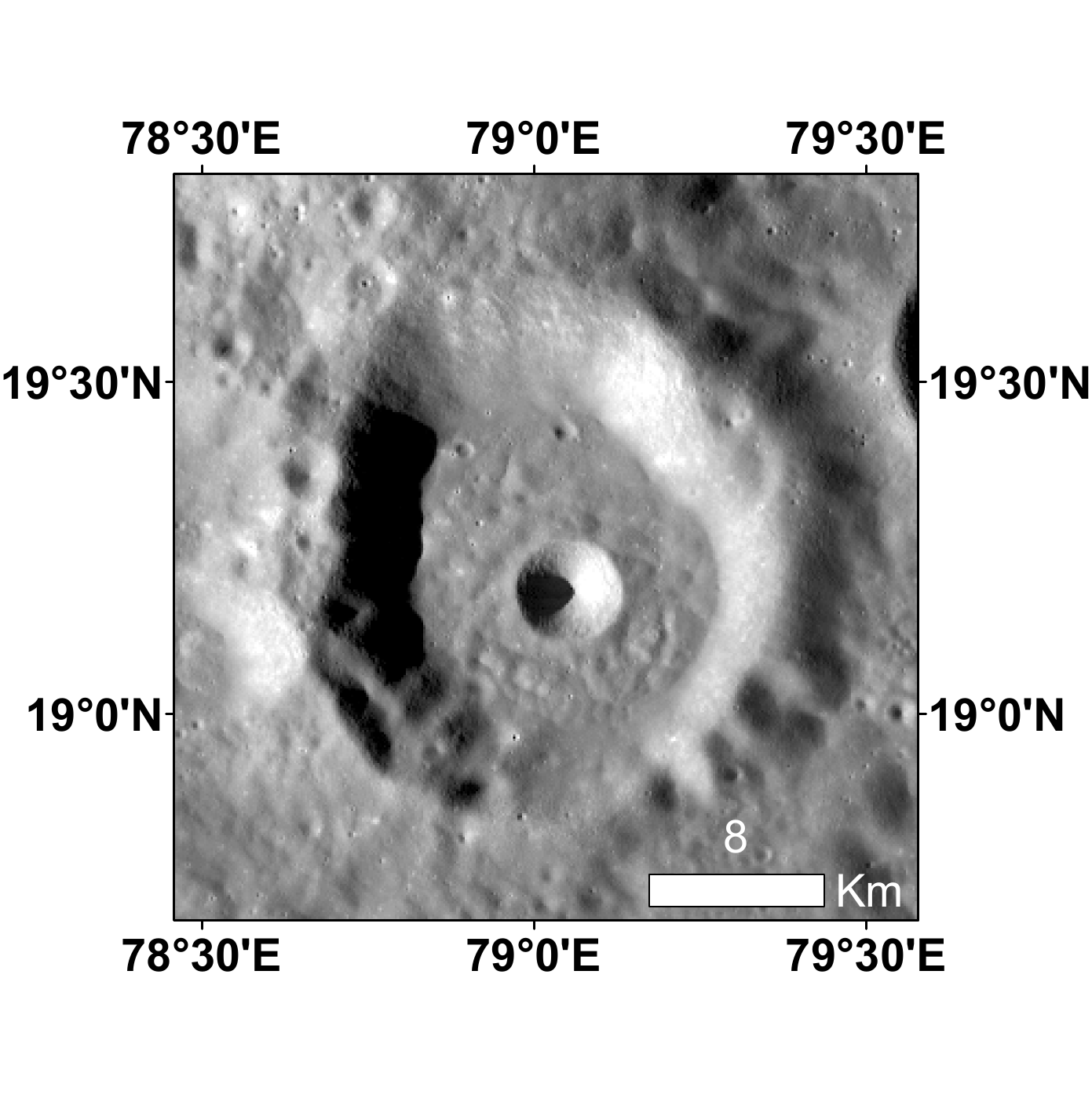}
         \caption{Crater within crater}
     \end{subfigure}     
        \caption{Visual inspection of size and shape variation of craters on the Luanr surface using LROC WAC Mosaic~\cite{robinson2010lunar}.}
        \label{fig:size_shape}
\end{figure*}

The planetary bodies will continue to be an important object of study because of recent advances in space exploration. 
Space exploration is used for many scientific studies with the primary goals of searching for evidence of water on the planetary bodies, understanding the origin of the Moon, Mars, or any other planetary bodies through mineral and chemical composition studies, mapping the planetary bodies' surface in greater detail and detecting and identifying the presence of atomic species in the planetary bodies' atmosphere~\cite{nazari2020water,hayne2015evidence,hargitai2019fundamental,waite2018chemical}.
As we expand our exploration of space and the solar system, the planetary bodies could serve as a repository for fuel, oxygen, and other critical raw materials~\cite{james2022meteorite}. 
For this purpose, craters are among the most studied geomorphic features because they can be utilized for studying past and present geological processes, the relative ages of craters, the energy source surrounding it, surface mineralogy, and chemical compositions~\cite{neukum2001cratering,solomonidou2020chemical,pan2017stratigraphy,ZHANG2021116896}. The crater's characteristics, such as variation in size, shape, floor structures, terrain properties, degree of degradation, and widely disparate distributions present several challenges for crater detection. Some of these challenges are described below:

\subsection{Size variation}
\label{subsec:Size variation}
Impact craters' size depends on various factors, such as the impactor's size, velocity, and surface properties of the impact site.
These factors lead to huge variability of craters' size from hundreds of meters to kilometers~\cite{pike1977size}.
Some of the craters with different sizes are shown in Figure~\ref{fig:size_shape}.
The wide range of craters' sizes makes it difficult to detect them all automatically. 
For example, if a low-resolution image is utilized for the automatic crater detection method, in this scenario, detecting small craters may not be possible since the number of pixels representing smaller craters will be significantly less. If a high-resolution image is utilized, then the detection of larger craters may not be possible due to limitations in computational power.

\subsection{Shape variation}
\label{subsec:Shape variation}

The crater shape can vary due to the impact angle, solar wind weathering, degradation, and differences in the crater-formation process. Most crater detection methods treat craters as circular shapes. However, as shown in Figure~\ref{fig:size_shape}, craters can be elliptical, irregular, or overlapping. Due to this, detecting all shapes of the craters using a single crater detection algorithm is challenging.

\subsection{Terrain variation}
\label{subsec:Terrain variation}
Different planetary bodies exhibit various surface properties. 
For example, the Moon's surface has highlands and lowlands, mountains, and volcanoes~\cite{mutch2015geology}, whereas the Martian surface is rocky with canyons, volcanoes, and dry lake beds, and most of its surface is covered with dust~\cite{martin2009mass,frey1979martian}. As a result, CDA trained on images from one planetary surface, such as the Moon, may not be as effective at detecting craters on another planetary surface, such as Mars. Additionally, the surface characteristics of planets may vary in terms of their location—for instance, the maria and highland areas on the Moon exhibit different surface properties. The maria are comparatively fresh areas on the Moon that resulted from huge impacts that pierced the Moon's crust and excavated basins~\cite{stuart1970lunar}. 
Later, volcanic eruptions brought liquid magma to the surface, filling the basins, which generated the big flat expanses we can still see once it cooled and solidified. 
As a result, the maria region has a flat surface with few craters. 
However, lunar highlands are older because they did not experience structural disruption due to volcanoes~\cite{wilhelms1987geologic}. As a result, they have a more complex surface, such as mountains, and contain more impact craters than the maria. 
Therefore, a single crater detection algorithm may not work on a different region of the planetary surface.

\subsection{Degree of degradation}
\label{subsec:Degree of degradation}
The variation in the degree of impact crater degradation can be used to analyze the surface properties and estimate the crater age. The processes such as weathering, lava flows, impacting, and downslope material movement can cause craters to erode continuously. To understand how a crater degradation process takes place in the top few meters of the regolith (surface material), they can be grouped into different classes. For example, It can be divided into three categories: fresh craters, moderately degraded, and highly degraded craters~\cite{agarwal2019study,marco2022depth}. Fresh craters are the least degraded, with sharp rims and bowl-shaped interiors; moderately degraded craters have rims that are lower than fresh craters but more bowl-shaped than highly degraded craters; highly degraded craters have no rims and shallow, funnel-shaped profiles~\cite{mahanti2018small}. These aid in determining the rate of crater degradation. Also, sometimes due to erosion, craters blend into the surface. 

These different degradation stages make it difficult for a single crater detection algorithm to detect all such craters.

\subsection{Different data-type}
\label{subsec:Different data-type}
The planetary data mainly used for crater detection are digital orthophoto maps (DOMs), digital elevation maps (DEMs), and near IR images. These data's characteristics differ from one another, such as DOMs and infrared images are affected by sun angle and cause highlight and shadow patterns. In contrast, DEMs are unaffected but lack complex terrain information~\cite{degirmenci2010impact,yang2019bayesian}. Due to differences in data type characteristics, CDAs trained on one type of data do not detect craters effectively on another data type. 
However, the unique characteristics of the different data lead to complementary information that can be useful for better CDA.
Hence, some research works, such as Tewari et al.~\cite{tewari2022automated} and Mao et al.~\cite{mao2022coupling}, used data fusion techniques to detect craters. 
However, existing methods that train CDAs on image fusion to improve the features available to the CNN are not applicable in all scenarios. For example, if we want to train CDA on another region using a data fusion technique, it is possible that some of the data types in a specific resolution may be unavailable, posing yet another challenge. 

\section{Categorization of Deep learning based method for crater detection}
\label{sec:Categorization of Deep learning based method for crater detection}

\begin{figure}[!ht]
	\includegraphics[width=\linewidth]{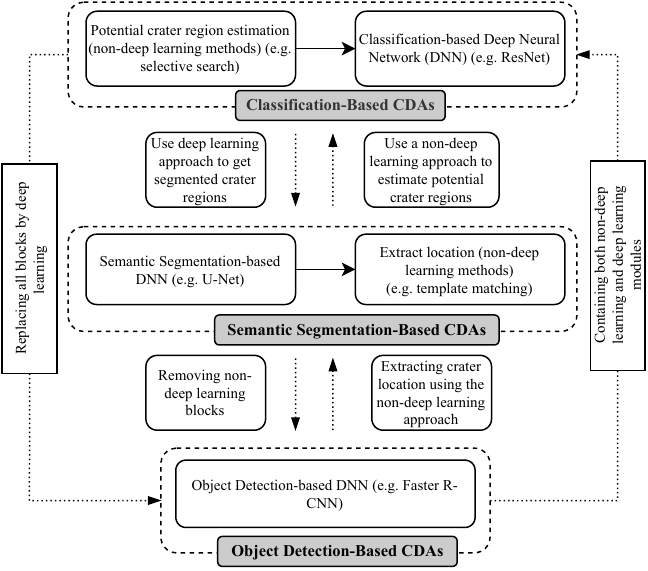}
	\caption{Categorization of Deep Learning based CDAs.}
	\label{fig:Survey_work_categorization}
\end{figure}

In computer vision, deep learning-based methods outperform traditional methods; therefore, recently, researchers followed the deep learning-based crater detection method for better generalization and performance. 
We divided the deep learning-based crater detection methods based on their approach into three categories: semantic segmentation-based, object detection-based, and classification-based (Figure~\ref{fig:Survey_work_categorization}). 
The details are provided in the following sections.
In addition, we have provided the highlights and properties of each CDA in Table~\ref{tab:cda_key_info}, which will assist researchers in understanding the key features of each CDA.

\begin{table*}[!ht]
\centering
\caption{Summary of highlights and properties of deep learning based CDAs.}
\label{tab:cda_key_info}
\resizebox{0.99\textwidth}{!}{%
\begin{tabular}{|lll|}
\hline
\multicolumn{1}{|l|}{\textbf{Author, Year}} &
  \multicolumn{1}{c|}{\textbf{Title}} &
  \multicolumn{1}{c|}{\textbf{Highlights and properties}} \\ \hline
\multicolumn{3}{|c|}{} \\
\multicolumn{3}{|c|}{\multirow{-2}{*}{Semantic Segmentation based Crater Detection}} \\ \hline
\multicolumn{1}{|l|}{\begin{tabular}[c]{@{}l@{}}Silburt et al.~\cite{silburt2019lunar}, \\ 2019\end{tabular}} &
  \multicolumn{1}{l|}{Lunar crater identification via deep learning} &
  \begin{tabular}[c]{@{}l@{}}Used U-Net framework~\cite{ronneberger2015u} for crater detection; Generalization capability checked \\ qualitatively on Mercury surface; First time  Head et al.~\cite{head2010global} and Povilaitis et al.~\cite{povilaitis2018crater}\\ catalog were utilized for training CDA; Detected new craters not listed in Head et al. \\ and Povilaitis et al..\end{tabular} \\ \hline
\multicolumn{1}{|l|}{\begin{tabular}[c]{@{}l@{}}Christopher Lee~\cite{lee2019automated},\\ 2019\end{tabular}} &
  \multicolumn{1}{l|}{Automated crater detection on Mars using deep learning} &
  \begin{tabular}[c]{@{}l@{}}Used U-Net framework for crater detection; Applied to the MOLA/HRSC blended\\ DTM on the Martian surface; Detected 75\% of the craters from the Robbins and\\ Hynek catalog~\cite{robbins2012new}.\end{tabular} \\ \hline
\multicolumn{1}{|l|}{\begin{tabular}[c]{@{}l@{}}DeLatte et al.~\cite{delatte2019segmentation},\\ 2019\end{tabular}} &
  \multicolumn{1}{l|}{\begin{tabular}[c]{@{}l@{}}Segmentation convolutional neural networks for automatic\\ crater detection on Mars\end{tabular}} &
  \begin{tabular}[c]{@{}l@{}}Inspired by U-Net, Crater U-Net is proposed; First paper to use Crater U-Net, a\\ segmentation CNN, to find Martian craters in THEMIS thermal infrared data;\\ Explore the effect of  parameters/hyperparameters such as kernel size, filters, and\\ amount of training data.\end{tabular} \\ \hline
\multicolumn{1}{|l|}{\begin{tabular}[c]{@{}l@{}}Wang S. et al.~\cite{wang2020effective},\\ 2020\end{tabular}} &
  \multicolumn{1}{l|}{\begin{tabular}[c]{@{}l@{}}An effective lunar crater recognition algorithm based on\\ convolutional neural network\end{tabular}} &
  \begin{tabular}[c]{@{}l@{}}Integrating the residual connection in U-Net; Achieved desirable detection results in\\ overlapping craters cases.\end{tabular} \\ \hline
\multicolumn{1}{|l|}{\begin{tabular}[c]{@{}l@{}}Lee et al.~\cite{lee2021automated},\\ 2021\end{tabular}} &
  \multicolumn{1}{l|}{Automated crater detection with human level performance} &
  \begin{tabular}[c]{@{}l@{}}ResUNet framework~\cite{zhang2018road} utilized to detect craters on optical imagery and digital \\ terrain model; The $F_1$-score of the proposed work is on par with the catalog \\ compared to another catalog or vice versa; Combine the detection of detected craters \\ from optical imagery and digital terrain model to improve the performance.\end{tabular} \\ \hline
\multicolumn{1}{|l|}{\begin{tabular}[c]{@{}l@{}}Jia et al.~\cite{jia2021moon},\\ 2021\end{tabular}} &
  \multicolumn{1}{l|}{\begin{tabular}[c]{@{}l@{}}Moon impact crater detection using nested attention\\ mechanism based UNet++\end{tabular}} &
  \begin{tabular}[c]{@{}l@{}}A new framework, nested attention aware U-Net (NAU-Net) is proposed, combining \\ UNet++~\cite{zhou2019unet++} and attention network~\cite{oktay2018attention}; It improves the efficiency of semantic \\ information propagation.\end{tabular} \\ \hline
\multicolumn{1}{|l|}{\begin{tabular}[c]{@{}l@{}}Chen et al.~\cite{chen2021lunar},\\ 2021\end{tabular}} &
  \multicolumn{1}{l|}{\begin{tabular}[c]{@{}l@{}}Lunar features detection for energy discovery via deep \\ learning\end{tabular}} &
  \begin{tabular}[c]{@{}l@{}}The first paper to detect craters and rilles to discover the potential energy sources; \\ Deploy the HRNET~\cite{sun2019deep} that can efficiently extract semantic and high-resolution \\ spatial information from input images.\end{tabular} \\ \hline
\multicolumn{1}{|l|}{\begin{tabular}[c]{@{}l@{}}Mao et al.~\cite{mao2022coupling},\\ 2022\end{tabular}} &
  \multicolumn{1}{l|}{\begin{tabular}[c]{@{}l@{}}Coupling complementary strategy to U-Net based\\ convolution neural network for detecting lunar impact\\ craters\end{tabular}} &
  \begin{tabular}[c]{@{}l@{}}A dual-path convolutional neural network proposed, which is based on a U-Net;\\ Utilizes the complementary information from elevation maps and optical images\end{tabular} \\ \hline
\multicolumn{3}{|c|}{} \\
\multicolumn{3}{|c|}{\multirow{-2}{*}{Object Detection based Crater Detection}} \\ \hline
\multicolumn{1}{|l|}{\begin{tabular}[c]{@{}l@{}}Ali-Dib et al.~\cite{ali2020automated},\\ 2020\end{tabular}} &
  \multicolumn{1}{l|}{\cellcolor[HTML]{FFFFFF}\begin{tabular}[c]{@{}l@{}}Automated crater shape retrieval using weakly-supervised \\ deep learning\end{tabular}} &
  \cellcolor[HTML]{FFFFFF}\begin{tabular}[c]{@{}l@{}}Mask R-CNN framework~\cite{he2017mask} is used to detect craters; Analyze crater ellipticity \\ and depth distribution using extracted shape from Mask R-CNN output.\end{tabular} \\ \hline
\multicolumn{1}{|l|}{\begin{tabular}[c]{@{}l@{}}Yang et al.~\cite{yang2020lunar},\\ 2020\end{tabular}} &
  \multicolumn{1}{l|}{\cellcolor[HTML]{FFFFFF}{\color[HTML]{222222} \begin{tabular}[c]{@{}l@{}}Lunar impact crater identification and age estimation with \\ Chang’E data by deep and transfer learning\end{tabular}}} &
  \cellcolor[HTML]{FFFFFF}\begin{tabular}[c]{@{}l@{}}R-FCN framework~\cite{dai2016r} is used to detect craters; IAU catalog is used for training; \\ Transfer learning-based approach follows for training R-FCN.\end{tabular} \\ \hline
\multicolumn{1}{|l|}{\begin{tabular}[c]{@{}l@{}}Hsu et al.~\cite{hsu2021knowledge},\\ 2021\end{tabular}} &
  \multicolumn{1}{l|}{\cellcolor[HTML]{FFFFFF}\begin{tabular}[c]{@{}l@{}}Knowledge-driven GeoAI: integrating spatial knowledge \\ into multi-scale deep learning for Mars crater detection\end{tabular}} &
  \cellcolor[HTML]{FFFFFF}\begin{tabular}[c]{@{}l@{}}A feature pyramid network is used for feature map generation of craters; Hough\\ transform is integrated into the deep learning process; Scale-aware object classifier\\ is used to improve the detection of smaller craters.\end{tabular} \\ \hline
\multicolumn{1}{|l|}{\begin{tabular}[c]{@{}l@{}}Zang et al.~\cite{zang2021semi},\\ 2021\end{tabular}} &
  \multicolumn{1}{l|}{\cellcolor[HTML]{FFFFFF}\begin{tabular}[c]{@{}l@{}}Semi-supervised deep learning for lunar crater detection\\ Using CE-2 DOM\end{tabular}} &
  \cellcolor[HTML]{FFFFFF}\begin{tabular}[c]{@{}l@{}}Two-teachers self-training with Noise (TTSN) is proposed to tackle incomplete \\ ground truth; Detection analysis was performed for maria and highland regions \\ of the lunar surface.\end{tabular} \\ \hline
\multicolumn{1}{|l|}{\begin{tabular}[c]{@{}l@{}}Jia Y. et al.~\cite{jia2021split},\\ 2021\end{tabular}} &
  \multicolumn{1}{l|}{\cellcolor[HTML]{FFFFFF}\begin{tabular}[c]{@{}l@{}}Split-attention networks with self-calibrated convolution \\ for Moon impact crater detection from multi-source data\end{tabular}} &
  \cellcolor[HTML]{FFFFFF}\begin{tabular}[c]{@{}l@{}}R-FCN~\cite{dai2016r} is utilized for crater detection; For feature extraction, inspired by \\ ResNest~\cite{zhang2020resnest} and the self-calibration convolution of SCNet~\cite{liu2020improving}, a \\ split-attention network with self-calibrated convolution (SCNeSt) is proposed; \\ Transfer learning based approach followed to detect craters on Mercury and Mars.\end{tabular} \\ \hline
\multicolumn{1}{|l|}{\begin{tabular}[c]{@{}l@{}}Yang et al.~\cite{yang2021high},\\ 2021\end{tabular}} &
  \multicolumn{1}{l|}{\cellcolor[HTML]{FFFFFF}\begin{tabular}[c]{@{}l@{}}High-resolution feature pyramid network for automatic \\ crater detection on Mars\end{tabular}} &
  \cellcolor[HTML]{FFFFFF}\begin{tabular}[c]{@{}l@{}}Emphasize on detection of small-scale craters; Adaptive anchor calculation and \\ label assignment algorithm (AACLA) is proposed to collect sufficient number \\ of small-scales craters for training; Proposed high-resolution feature pyramid \\ network (HRFPNet) with feature aggregation module and balanced regression loss.\end{tabular} \\ \hline
\multicolumn{1}{|l|}{\begin{tabular}[c]{@{}l@{}}Yang H et al.~\cite{yang2021craterdanet},\\ 2021\end{tabular}} &
  \multicolumn{1}{l|}{\cellcolor[HTML]{FFFFFF}\begin{tabular}[c]{@{}l@{}}CraterDANet: A convolutional neural network for small-\\ -scale crater detection via synthetic-to-real domain \\ adaptation\end{tabular}} &
  \cellcolor[HTML]{FFFFFF}\begin{tabular}[c]{@{}l@{}}Domain adaptation technique is used to detect craters on the Moon; Inspired by \\ CycleGAN~\cite{zhu2017unpaired}, cycle consistency loss is used to achieve feature level distribution \\ alignment in crater detection; Present a lunar crater dataset of small craters \\ containing 20000 craters.\end{tabular} \\ \hline
\multicolumn{1}{|l|}{\begin{tabular}[c]{@{}l@{}}Lin et al.~\cite{lin2022lunar},\\ 2022\end{tabular}} &
  \multicolumn{1}{l|}{\begin{tabular}[c]{@{}l@{}}Lunar crater detection on digital elevation model: a \\ complete workflow using deep learning and its \\ application\end{tabular}} &
  \begin{tabular}[c]{@{}l@{}}Exhaustive anlaysis of 9 object detection architectures is done; Crater validation \\ tool developed for manual verification of craters; Multiscale grid cropping \\ approach used for data generation\end{tabular} \\ \hline
\multicolumn{3}{|c|}{} \\
\multicolumn{3}{|c|}{\multirow{-2}{*}{Classification based Crater Detection}} \\ \hline
\multicolumn{1}{|l|}{\begin{tabular}[c]{@{}l@{}}Emami et al.~\cite{emami2019crater},\\ 2019\end{tabular}} &
  \multicolumn{1}{l|}{\begin{tabular}[c]{@{}l@{}}Crater detection using unsupervised algorithms and \\ convolutional neural networks\end{tabular}} &
  \begin{tabular}[c]{@{}l@{}}Four unsupervised algorithms, i.e.,  Hough transform, highlight-shadow regions, \\ convex  grouping, and interest points investigated; CNN is utilized for classification \\ of crater and non-crater; Interest point or convex grouping with the CNN \\ classification network is the most optimistic crater detection approach; Detect small \\ size craters of diameter range 20 to 200 meters.\end{tabular} \\ \hline
\end{tabular}%
}
\end{table*}

\subsection{Semantic Segmentation Based CDAs}
\label{sec:Semantic Segmentation Based Method}

\begin{figure*}[!ht]
	\includegraphics[width=0.95\linewidth]{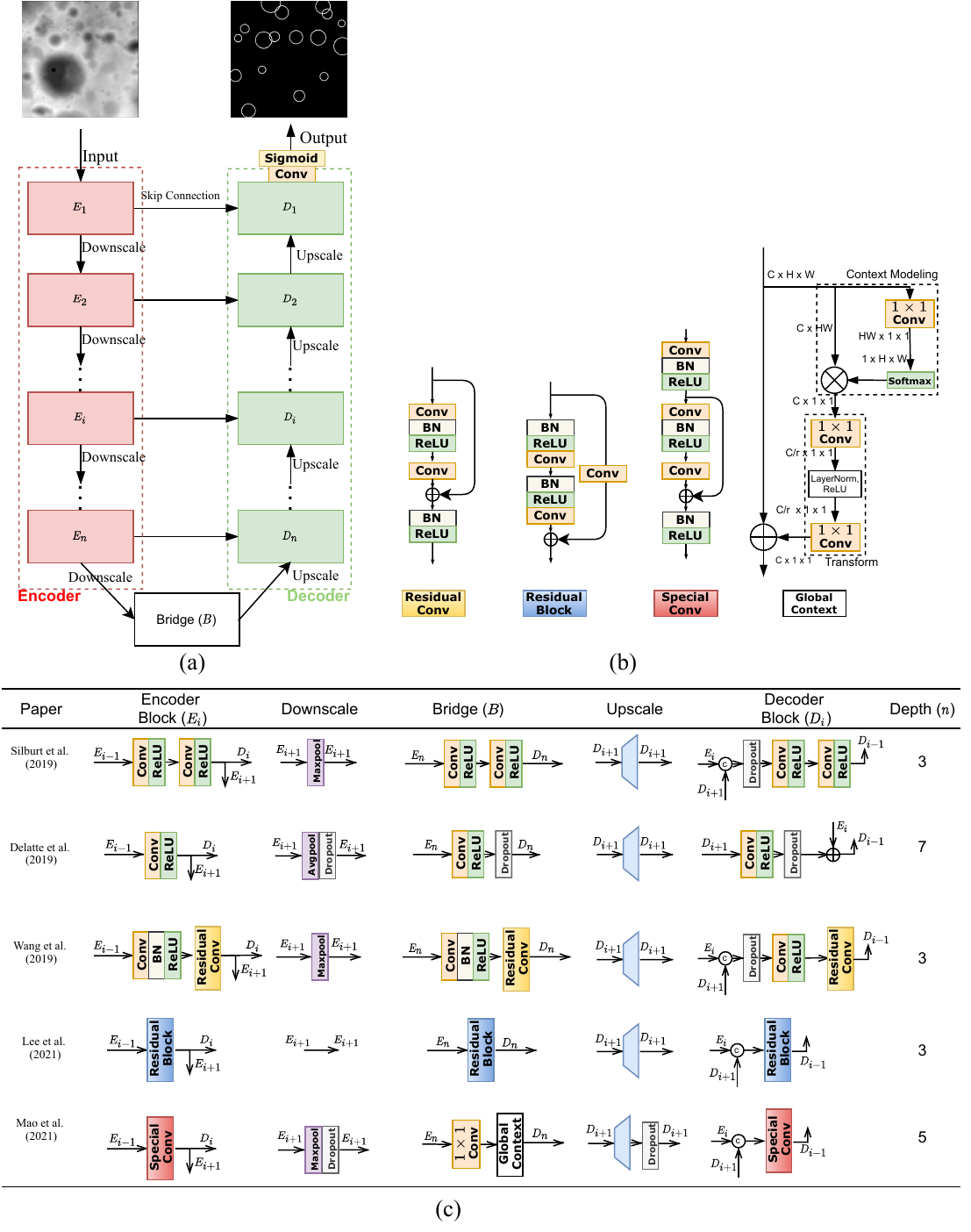}
	\caption{Overview of Semantic Segmentation based Crater Detection Algorithms. (a) Encoder-bridge-decoder structure. (b) Details of modules mentioned in (c). (c) Encode-bridge-decoder framework in existing works.}
	\label{fig:sementic_seg_all}
\end{figure*}

A semantic segmentation network is used to categorize each pixel in the image using predefined labels. 
For crater detection, each pixel is categorized as crater or non-crater. Since the network does not provide the location and size information of craters, the output of the segmentation network is further needed to process.
It can be accomplished with the help of techniques such as the Hough transform and template matching. However, the majority of the semantic segmentation-based crater detection work employed a template matching algorithm. Following their footsteps, we also utilized the template matching algorithms in our work.

Most of the semantic segmentation-based CDAs followed the encoder-bridge-decoder structure. Research works~\cite{silburt2019lunar,delatte2019segmentation,wang2020effective,lee2019automated,lee2021automated,mao2022coupling} are explained using encoder-bridge-decoder structure shown in Figure~\ref{fig:sementic_seg_all}. 
The encoder and decoder network consist of multiple encoder and decoder blocks. 
The $ith$ encoder block $E_i$ received input from the downsampled output of $E_{i-1}$, and its outputs were passed to the $D_i$ and $E_{i+1}$. 
Similarly, the $ith$ decoder block $D_i$ received input from the $E_i$ and an upsampled version of the $D_{i+1}$, and its output was sent to the $D_{i-1}$. The number of encoder or decoder blocks determines the depth ($n$) of the architecture. 
As shown in Figure~\ref{fig:sementic_seg_all}(a), the encoder network is on the left side of the figure. 
It performs the feature extractor function by using a series of encoder blocks $[E_{1}, E_{2}, E_{3}, ..., E_{i}, ..., E_{n}]$ and learns abstract patterns from the input image. The encoder network reduces the spatial dimension of features while increasing the number of feature channels to obtain high semantic information. 
  
A downscale operation is performed after every encoder block to reduce the spatial dimension of the input and learn the feature representation at multiple levels.
In existing CDAS, It comprises either a 2$\times$2 max pooling layer with a dropout, an average pooling with a dropout, or a 2$\times$2 max pooling layer.
Similarly, the decoder network is present on the right side of the architecture. It consist of series of decoder blocks $[D_{n}, D_{n-1}, D_{n-2}, ...,D_{i}, ...,D_{1}]$. 
The decoder network aids in increasing the spatial dimension of the features while decreasing the number of feature channels used to extract the high-dimensional feature representation for better localization.

An upscale operation is performed after each decoder block using UpSampling2D or Conv2DTranspose available at TensorFlow Keras. UpSampling2D, by default, does nearest neighbor interpolation, whereas Conv2DTranspose is a convolution operation that learns up-sample features.
Research works~\cite{silburt2019lunar, delatte2019segmentation, lee2021automated} used UpSampling2D whereas~\cite{wang2020effective, mao2022coupling}  used Conv2DTranspose to perform the upscale operation.
Each encoder block ($E_i$) feature is passed to the respective decoder block ($D_i$) using a skip connection. These skip connections efficiently transfer spatial and semantic information throughout the network by improving gradient flow during back-propagation and allowing the network to learn better representation. The bridge ($B$) connects the last encoder and first decoder block, completing the information flow. Finally, the segmented mask of craters is produced from the final decoder $D_{1}$ output after it has undergone a $1\times1$ convolution with a sigmoid activation function.    

Few researchers borrowed the idea of a residual connection from He et al.~\cite{he2016deep} and proposed a residual block named Residual Conv~\cite{wang2020effective}, Residual Block~\cite{lee2021automated}, and Special Conv~\cite{mao2022coupling} are shown in Figure~\ref{fig:sementic_seg_all}(b).
The detail of each architecture encoder block, decoder block, bridge, downscale, and upscale information is provided in Figure~\ref{fig:sementic_seg_all}(c). 
In Figure~\ref{fig:sementic_seg_all}(b) and Figure~\ref{fig:sementic_seg_all}(c), `Conv' define a convolution layer with kernel size of $3\times3$, `BN' define a batch normalization operation and `ReLU' define a rectified linear unit~\cite{nair2010rectified} activation function.

Many researchers, including those working on crater detection and classification domain, were intrigued when the U-Net architecture~\cite{ronneberger2015u} was proposed.
As a result, Silburt et al.~\cite{silburt2019lunar}, DeLatte et al.~\cite{delatte2019segmentation}, and Lee et al.~\cite{lee2019automated} used the U-Net architecture to detect the craters. However, their architectures differ from the original U-Net architecture~\cite{ronneberger2015u} primarily in terms of the number of filters, kernel size, and depth. 

The U-Net is inefficient in fusing multi-scale information and fails to adequately explore the high-resolution information from the input image effectively. 
Hence, we require an architecture that can enhance the learning of various scale features and strengthen feature transmission across the network so that small and overlapping craters can easily be detected with high accuracy. As a result, researchers began attempting to create numerous U-Net variants in order to obtain more accurate segmentation results.

For example, Wang et al.~\cite{wang2020effective} proposed an effective residual U-Net (ERU-Net) architecture for crater detection by replacing a convolution block in U-Net with a residual block named ``Residual Conv'' and it is shown in Figure~\ref{fig:sementic_seg_all}(b) and Figure~\ref{fig:sementic_seg_all}(c). They were inspired by He et al.~\cite{he2016deep}, who proposed a deep residual framework that used identity mapping by shortcut connection to ease the training and overcome the degeneration issue in deep neural network training. Similarly, Lee et al.~\cite{lee2021automated} used the ResUnet~\cite{zhang2018road} architecture, which also uses residual connections in U-Net to enhance the network's learning ability. The residual block used in the architecture is named ``Residual Block'' and it is shown in  Figure~\ref{fig:sementic_seg_all}(b) and Figure~\ref{fig:sementic_seg_all}(c).

As discussed in Section~\ref{sec:Challenges for crater detection}, DEM and optical image each have their own set of benefits and drawbacks, and using only one type of data may limit the insufficient crater feature extraction.
Therefore to utilize the complimentary information of DEMs and optical images, Mao et al.~\cite{mao2022coupling} proposed a dual path convolutions neural network that integrates the features of DEMs and optical images. Similar to other CDAs, i.e., Wang et al.~\cite{wang2020effective} and Lee et al.~\cite{lee2021automated}, it also replaces the plain convolution block in U-Net with a residual block named `Special Conv' in Figure~\ref{fig:sementic_seg_all}(b) and Figure~\ref{fig:sementic_seg_all}(c).
In the encoder network, DEM and optical image features are extracted independently. The extracted features of DEM and optical images in the bridge network are integrated. Finally, the decoder network with the attention mechanism enlarges the features to optimize the feature information further and obtain the segmented output.

\begin{figure}[!ht]
	\includegraphics[width=1\linewidth]{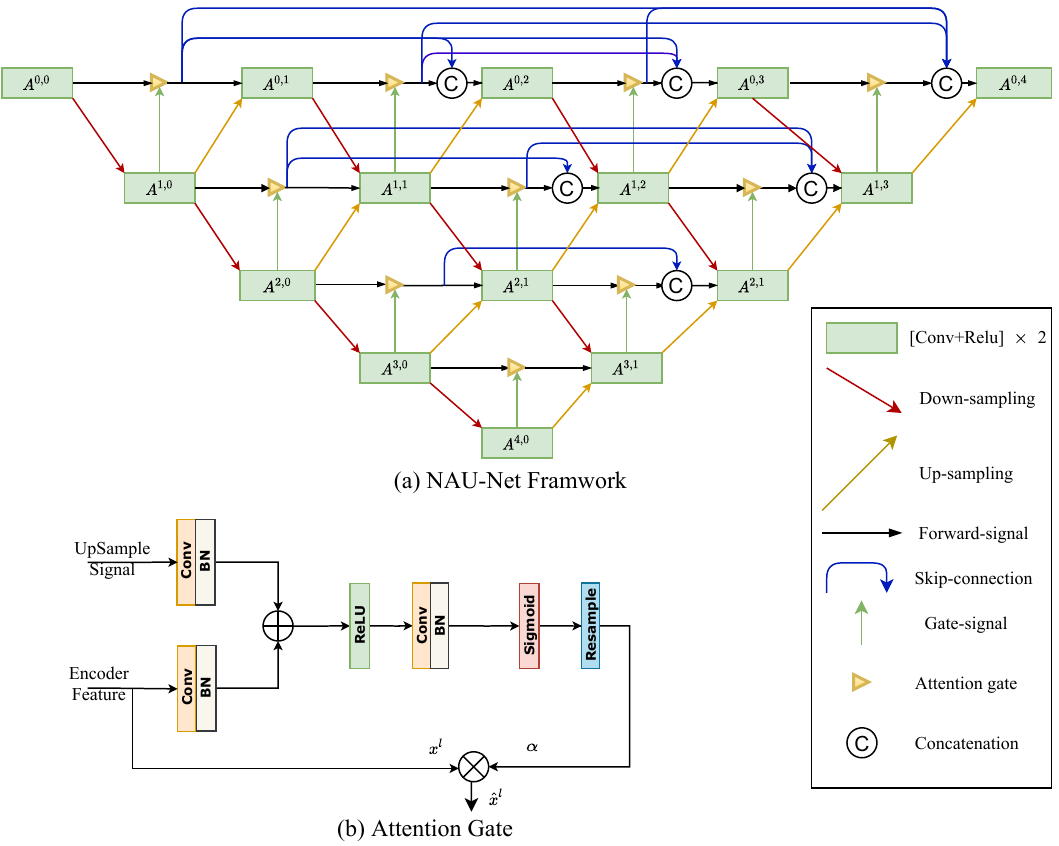}
	\caption{Network architecture of NAU-Net~\cite{jia2021moon}.}
	\label{fig:NAUNet}
\end{figure}
In the general encoder-decoder architecture, each encoder block is connected to the decoder block with a skip connection to reduce the loss of high-level features due to downsampling in the encoder. However, in Jia et al.~\cite{jia2021moon}, the skip connections are replaced by nested dense connections to better preserve high-level features for detecting smaller craters. The author proposed another variant of the U-Net architecture called NAU-Net, which combines the U-Net~\cite{ronneberger2015u} and attention gates~\cite{oktay2018attention} with a nested dense connection (Figure~\ref{fig:NAUNet}). The attention gate used in the architecture helps to improve feature extraction ability in order to detect overlapping craters.

\begin{figure}[!ht]
	\includegraphics[width=1\linewidth]{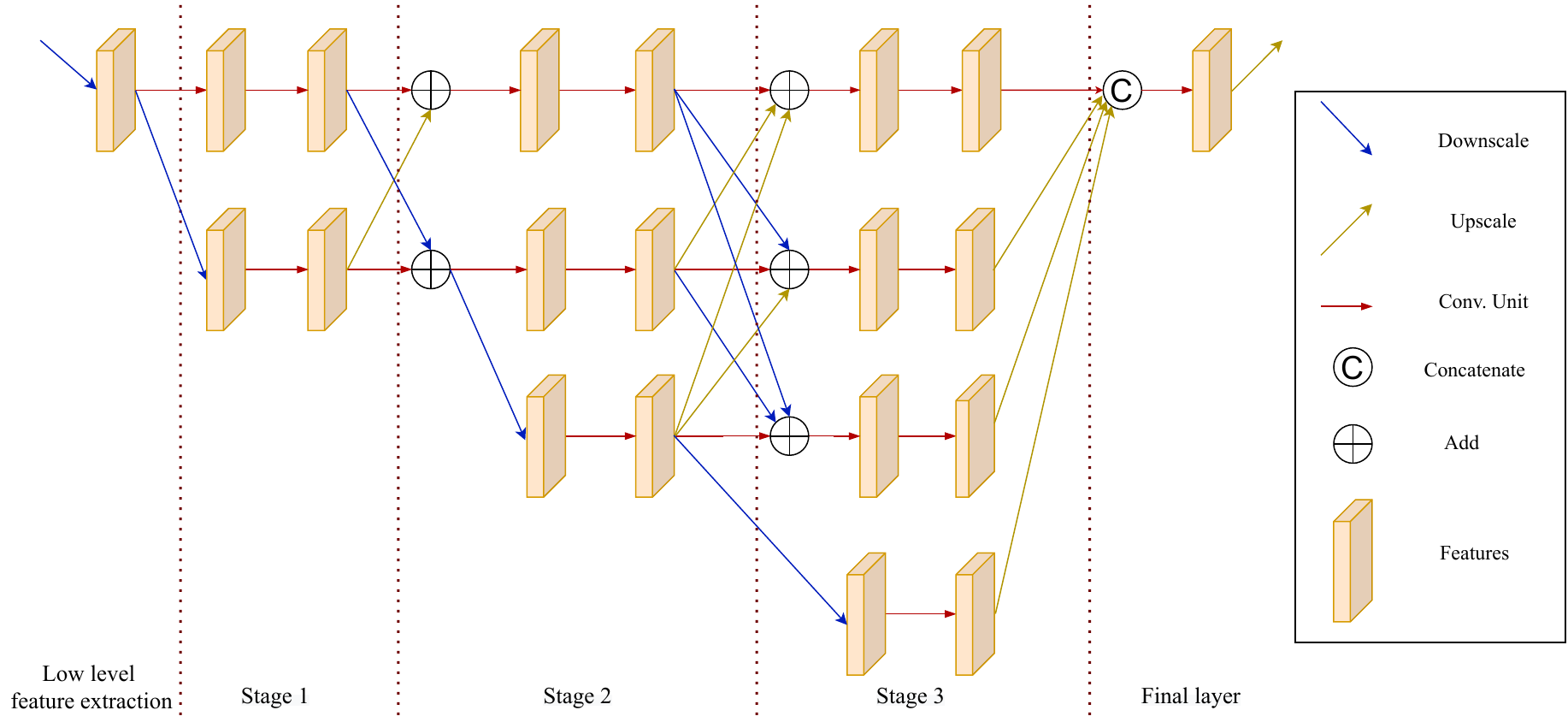}
	\caption{Overview of HRNet framework~\cite{sun2019deep}.}
	\label{fig:HRNet}
\end{figure}

Chen et al.~\cite{chen2021lunar} used the HRNet~\cite{sun2019deep} framework (Figure~\ref{fig:HRNet}) to detect craters and rilles on the lunar surface. To detect craters on the lunar surface, the HRNet was first trained on the large dataset generated by Silburt et al.~\cite{silburt2019lunar}, then fine-tuned on the smaller manually annotated dataset containing $44$ images of the lunar surface.
HRNet has overcome the limitations of the U-Net framework by preserving the high-resolution input data information in deeper layers and learning the comprehensive representation of input by multi-scale fusion. 
It consists of four stages, which have multiple residual blocks~\cite{he2016deep} to extract features of different sizes. After each stage: the features are down-sampled by a factor of two, the number of channels is increased, and multiple up-sampling and down-sampling operations are performed for multi-scale fusion.

\subsubsection{Limitations}
\label{subsubsec:Limitations}

The main issue with the semantic segmentation-based CDAs is that they detect craters in two asynchronous steps. 
The semantic segmentation framework is GPU-based, and location and size extraction methods, such as template matching, are CPU based, resulting in an imbalanced runtime comparison.
The template matching algorithm is slow, which causes high time complexity, and a fixed threshold in template matching may not be optimal for complex and incomplete segmentation.

\subsubsection{Implementation}
\label{subsubsec:Implementation}
$\newline$
\vspace{-4mm}

\paragraph{Dataset Preparation}
We have utilized the DEM data from Tewari et al.~\cite{tewari2022automated}, having a resolution of $100$ m/pixel.
The DEM mosaic was obtained through LRO's lunar orbiter laser altimeter (LOLA) and the SELENE terrain camera (TC) (\cite{bib:barker2016new}). In the data generation process, DEMs of size $1024 \times 1024$ are cropped from the mosaic using a raster method, with adjacent images overlapping by $50\%$. The $50\%$ overlapping strategy used while cropping the image ensures that many of the craters cropped in that image will have a chance to appear in the adjacent image. 
The cropped DEMs are resized to $512 \times 512$ pixels. The provided DEMs have a simple cylindrical projection. In our work, we have converted it into an orthographic projection to reduce image distortion.

For each DEM, we have constructed the target image of size $512\times512$ pixels. The target image is a binary image, where the crater rim is marked as a ring with a thickness of  $2$ pixels. The craters centers and radius from the catalog~\cite{povilaitis2018crater} are used for marking craters in the target image.
Povilaitis et al.\cite{povilaitis2018crater} catalog contains craters of diameter size $5$ to $20$ km. This catalog is conservative since it only includes craters that are highly certain to be a crater. 
The study area spans longitude from -$180^{\circ}$ to $180^{\circ}$ and latitude $\pm 60^{\circ}$. The training region contains longitude from -$180^{\circ}$ to $60^{\circ}$ and latitude $\pm 60^{\circ}$, and the testing region spans longitude range from $60^{\circ}$ to $180^{\circ}$ and latitude $\pm 60^{\circ}$. 
The total number of training, validation, and testing patches are $6623$, $1520$, and $5041$, respectively, with $9191$, $2379$, and $7765$ craters.

\paragraph{Setting}

In our experiment, learning rate=$0.001$, batch size=$2$, epochs=$50$, and Adam optimizer~\cite{kingma2014adam} is used. The loss function is binary cross entropy. 
The model with the lowest validation loss is chosen to predict craters. 
The data, code, and extracted craters location information will be provided in the following link~\footnote{\href{https://github.com/Ataltewari/Review-Work}{https://github.com/Ataltewari/Review-Work}}. 
We utilized the open-source code of the following papers:~\cite{silburt2019lunar,delatte2019segmentation,lee2019automated,lee2021automated,chen2021lunar}, for training and testing on our generated dataset.
For training and testing the Jia et al.~\cite{jia2021moon} crater detection method, code from the following link~\footnote{\url{https://github.com/rizalmaulanaa/Robustness_of_Prob_U_Net}} is utilized. 
As the code is unavailable in Mao et al.~\cite{mao2022coupling} and Wang et al.~\cite{wang2020effective} paper, we implemented it using the information provided in their respective papers. 
Mao et al.~\cite{mao2022coupling} proposed framework utilized both optical and DEM data for training the deep neural network; however, in our work, we have only used DEM to maintain data type uniformity with other frameworks.

\paragraph{Crater extraction}

The segmentation network output is further processed to get the craters' location and size. 
The segmented output value varies between $0$ to $1$. It is converted to a binary image with a threshold, $B$. If the value is greater than $B$, it is set to $1$; otherwise, it is set to $0$.
The binary image is used at the input of the template matching algorithm to extract the craters' location and size information.
The template matching algorithm's radius range varies from $r_{min}$ to $r_{max}$. 
The function of the algorithm is to calculate a match probability for circles of radius varying from $r_{min}$ to $r_{max}$ for each pixel in an image. If the probability value is above a threshold $Pm$, that ring is marked as a crater.

Following Silburt et al.~\cite{silburt2019lunar}, duplicate removal is done on a pixel scale for craters detected in each DEM using the following equations:
\begin{align}
    \frac{(x_i - x_j)^2+(y_i - y_j)^2}{\text{min}(r_i,r_j)^2} & < D_{x,y}\\
    \frac{|r_i - r_j|}{\text{min}(r_i,r_j)} & < D_r
\end{align}
where ($x_i,y_i$) is the center location and $r_i$ is the radius of the $ith$ crater in pixels. 
$D_{x,y}$ and $D_r$ are tunable hyperparameters. |a| represent the absolute value of a.

The predictions with maximum template matching probability are retained during duplicate removal for all the architectures. 
We evaluated post-CNN metrics on the validation data in the Silburt et al.~\cite{silburt2019lunar}, over the following hyperparameters range:
\begin{align*}
    B&=[0.01,0.05,0.1,0.15]\\
    Pm&=[0.3,0.4,0.5,0.6,0.7]\\
    D_{x,y}&=[0.6, 0.8, 1.0, 1.2, 1.4, 1.6, 1.8, 2.0, 2.2]\\
    D_r&=[0.2, 0.4, 0.6, 0.8, 1.0, 1.2]    
\end{align*}

The best $F_2$-score was obtained for $B=0.1$, $Pm=0.4$, $D_{x,y}=1.8$, $D_r=1.6$. 

\paragraph{Post processing}
The dataset contains images with $50\%$ overlap; each crater may appear in two or three images. It improves the probability of detecting craters, but the same crater may be detected in multiple images. Such duplicate detection is undesirable and needs to be filtered. Therefore we convert the pixel coordinates to degrees and kilometers coordinates and remove the duplicate craters.
The conversion is done by the following equations as provided in DeepMoon~\cite{silburt2019lunar}:
\begin{align}
    La-La_c &= \frac{\Delta L}{\Delta H} (y-y_c)\\
    Lo-Lo_c &= \frac{\Delta L}{\cos{(\frac{\pi La}{180^\circ})}\Delta H} (x-x_c)\\
    R &= r \frac{\Delta L}{C_{KD} \Delta H}\\
    C_{KD} &= \frac{180^\circ}{\pi R_{moon}}
\end{align}
where, in pixel coordinates, (x,y) is the central location and r is the radius.
In degree and kilometer coordinates,  $L_o$ and $L_a$ are longitude and latitude in degree, and $R$ is the radius in km. 
Subscript $c$ defines the center of the DEM. $\Delta H$ and $\Delta L$ are the pixel and latitude extent of the DEM. $C_{KD}$ is the kilometer to degree conversion factor, and $R_{moon}$ is the moon's radius in km.

Now, we removed the duplicate craters that satisfy the following equations,
\begin{align}
    \frac{(Lo_i-Lo_j)^2\cos^2{(\frac{\pi}{180^\circ}}\langle La\rangle)+ (La_i - La_j)^2}{C_{KD}^2 \text{min}(R_i,R_j)^2} &< D_{Lo,La}\\
    \frac{|R_i - R_j|}{\text{min}(R_i,R_j)}  &< D_R
\end{align}

where, $\langle La\rangle$=$\frac{1}{2}(La_i + La_j)$. 
$D_{Lo,La}$ and $D_R$ are post processing hyperparameters that are tuned on validation set, for the following range:
\begin{align*}
    D_{La,Lo}&=[0.2, 0.6, 1.0, 1.4, 1.8, 2.2, 2.6, 3.0, 3.4, 3.8, 4.2]\\
    D_R&=[0.2, 0.6, 1.0, 1.4, 1.8, 2.2, 2.6, 3.0, 3.4, 3.8, 4.2]
\end{align*}

\paragraph{Results and Discussion}

All existing semantic segmentation-based CDA models were trained and tested under identical conditions, such as the same training and testing region, data type, and diameter range.  The comparison results of all CDA models are shown in Table~\ref{tab:implementation}. 
The following metrics are used to evaluate the CDA's performance: precision, recall, $F_1$-score, $F_2$-score, median fractional latitude error, median fractional longitude error, and median fractional radial error.
In addition, we compared the training and inference time, total training parameters, model size, and total detection.

\begin{table*}[!ht]
\centering
\caption{Performance of different semantic segmentation-based CDAs}
\label{tab:implementation}
\resizebox{\textwidth}{!}{%
\begin{tabular}{lcccccccccccc} \hline
\textbf{Existing Works} &
  \textbf{\begin{tabular}[c]{@{}c@{}}Precision\\ (\%)\end{tabular}} &
  \textbf{\begin{tabular}[c]{@{}c@{}}Recall\\ (\%)\end{tabular}} &
  \textbf{\begin{tabular}[c]{@{}c@{}}$F_1$- \\ Score\\ (\%)\end{tabular}} &
  \textbf{\begin{tabular}[c]{@{}c@{}}$F_2$- \\ Score\\ (\%)\end{tabular}} &
  \textbf{\begin{tabular}[c]{@{}c@{}}Median\\ Latitudinal\\ Error\end{tabular}} &
  \textbf{\begin{tabular}[c]{@{}c@{}}Median\\ Longitudinal\\ Error\end{tabular}} &
  \textbf{\begin{tabular}[c]{@{}c@{}}Median\\ Radial\\ Error\end{tabular}} &
  \textbf{\begin{tabular}[c]{@{}c@{}}Training \\ Parameters (M)\end{tabular}} &
  \textbf{\begin{tabular}[c]{@{}c@{}}Training\\ Time (min/epoch)\end{tabular}} &
  \textbf{\begin{tabular}[c]{@{}c@{}}Inference\\ Time (ms)\end{tabular}} &
  \textbf{\begin{tabular}[c]{@{}c@{}}Model\\ Size (MB)\end{tabular}} &
  \textbf{\begin{tabular}[c]{@{}c@{}}Total Detected\\ Craters\end{tabular}} \\ \hline
Silburt et al.~\cite{silburt2019lunar} &
  62.46 &
  92.96 &
  74.71 &
  84.68 &
  4.62 &
  6.12 &
  4.13 &
  10.28 &
  {\color[HTML]{343434} 39} &
  75.06 &
  123.4 &
  \textbf{12,160} \\
DeLatte et al.~\cite{delatte2019segmentation} &
  64.56 &
  91.27 &
  75.63 &
  84.30 &
  4.80 &
  6.34 &
  4.23 &
  \textbf{0.73} &
  \textbf{2.37} &
  \textbf{11.34} &
  \textbf{8.9} &
  11,521 \\
Wang et al.~\cite{wang2020effective} &
  \textbf{70.89} &
  91.31 &
  \textbf{79.82} &
  \textbf{86.33} &
  \textbf{4.49} &
  6.00 &
  4.44 &
  23.74 &
  73.88 &
  136.73 &
  285.2 &
  10,480 \\
Lee et al.~\cite{lee2021automated} &
  66.92 &
  90.05 &
  76.78 &
  84.22 &
  4.60 &
  \textbf{5.99} &
  5.06 &
  8.30 &
  18.71 &
  29.27 &
  99.9 &
  10,903 \\
Mao et al.~\cite{mao2022coupling} &
  63.76 &
  92.92 &
  75.62 &
  85.13 &
  4.69 &
  6.06 &
  4.51 &
  10.03 &
  13.16 &
  32.47 &
  120.9 &
  11,855 \\
Jia et al.~\cite{jia2021moon} &
  70.14 &
  90.20 &
  78.91 &
  85.32 &
  5.01 &
  6.19 &
  4.33 &
  11.98 &
  46.91 &
  91.9 &
  144.5 &
  10,334 \\
Chen et al.~\cite{chen2021lunar} &
  65.63 &
  \textbf{93.15} &
  77.00 &
  85.94 &
  4.79 &
  6.06 &
  \textbf{3.99} &
  9.52 &
  46.82 &
  67.25 &
  115.4 &
  11,345 \\ \hline
\end{tabular}%
}
\end{table*}

Table~\ref{tab:implementation} shows that Wang et al.~\cite{wang2020effective} (ERU-Net) has the best precision ($70.89\%$), $F_1$-score ($79.82\%$), and $F_2$-score ($86.33\%$) compared to other works. 
Also, recall ($91.31\%$) is better than Jia et al.~\cite{jia2021moon}, DeLatte et al.~\cite{delatte2019segmentation}, and Lee et al.~\cite{lee2021automated}, indicating reliable detection of most impact craters. 
However, the training and inference time, training parameters, and memory acquired by the model are very high compared to other CDAs. 
Whereas Jia et al.~\cite{jia2021moon} has less space-time complexity, its precision and $F_1$-score are comparable to Wang et al.
Therefore, Jia et al.~\cite{jia2021moon} will be a better option than Wang et al.~\cite{wang2020effective} for applications requiring high precision and limited computational resource with less time complexity.  

Chen et al.~\cite{chen2021lunar} (HRNet) has the best recall ($93.15\%$); it may be due to the preservation of high-level features.
Silburt et al.~\cite{silburt2019lunar} and Mao et al.~\cite{mao2022coupling} have comparable recall with respect to Chen et al.~\cite{chen2021lunar}. 
However, Chen et al.~\cite{chen2021lunar} also has less number of parameters and model size compared to Silburt et al.~\cite{silburt2019lunar} and Mao et al.~\cite{mao2022coupling}. Hence, Chen et al.~\cite{chen2021lunar} is the best option for applications that require high recall, such as crater counting and hazard detection.

DeLatte et al.~\cite{delatte2019segmentation} has the lowest space-time complexity. 
It has $\sim$ $11$ times fewer parameters and model size than the other CDAs.
In addition, the recall is better than~\cite{lee2021automated} (ResUnet),~\cite{jia2021moon} (NAU-Net), and comparable with~\cite{wang2020effective} (ERU-Net). 
One interesting fact is that the recall of DeLatte et al.~\cite{delatte2019segmentation} is better than that of much more complex architectures such as NAUNet~\cite{jia2021moon} and ResUnet~\cite{lee2021automated}, which utilize residual and dense connections. 
DeLatte et al. will be the best choice for applications requiring both accuracy and space-time complexity, such as spacecraft landing.

Wang et al.~\cite{wang2020effective} and Lee et al.~\cite{lee2021automated} replace standard convolution with residual convolution; however, the recall is suboptimal. It signifies that may be replacing standard convolution with residual convolution is not sufficient to extract all the craters.

\begin{table*}[!ht]
\centering
\caption{Performance after ensemble different CDAs.}
\label{tab:imp_combine}
\resizebox{0.85\textwidth}{!}{%
\begin{tabular}{
>{\columncolor[HTML]{FFFFFF}}l 
>{\columncolor[HTML]{FFFFFF}}c 
>{\columncolor[HTML]{FFFFFF}}c 
>{\columncolor[HTML]{FFFFFF}}c 
>{\columncolor[HTML]{FFFFFF}}c 
>{\columncolor[HTML]{FFFFFF}}c 
>{\columncolor[HTML]{FFFFFF}}c 
>{\columncolor[HTML]{FFFFFF}}c }
\hline
\cellcolor[HTML]{FFFFFF} &
  \cellcolor[HTML]{FFFFFF} &
  \cellcolor[HTML]{FFFFFF} &
  \cellcolor[HTML]{FFFFFF} &
  \cellcolor[HTML]{FFFFFF} &
  \multicolumn{3}{c}{\cellcolor[HTML]{FFFFFF}{Total Detected Craters}} \\ \cline{6-8}
\multirow{-2}{*}{\cellcolor[HTML]{FFFFFF}CDAs Detection} &
  \multirow{-2}{*}{\cellcolor[HTML]{FFFFFF}Precision (\%)} &
  \multirow{-2}{*}{\cellcolor[HTML]{FFFFFF}Recall (\%)} &
  \multirow{-2}{*}{\cellcolor[HTML]{FFFFFF}$F_1$-Score (\%)} &
  \multirow{-2}{*}{\cellcolor[HTML]{FFFFFF}$F_2$-Score (\%)} &
  \begin{tabular}[c]{@{}c@{}}In \\ Ground-truth\end{tabular} &
  \begin{tabular}[c]{@{}c@{}}Not in \\ Ground-truth\end{tabular} &
  Total \\ \hline
All CDAs             & 51.73 & 97.18 & 67.52 & 82.65 & 7546 & 7041 & 14587 \\
\textgreater{}1 CDAs & 58.88 & 95.93 & 72.97 & 85.20 & 7449 & 5202 & 12651 \\
\textgreater{}2 CDAs & 63.63 & 94.84 & 76.16 & 86.36 & 7364 & 4209 & 11573 \\
\textgreater{}3 CDAs & 67.68 & 93.61 & 78.56 & 86.95 & 7269 & 3471 & 10740 \\
\textgreater{}4 CDAs & 71.53 & 91.65 & 80.35 & 86.77 & 7117 & 2833 & 9950  \\
\textgreater{}5 CDAs & 75.46 & 89.22 & 81.77 & 86.08 & 6928 & 2253 & 9181  \\
\textgreater{}6 CDAs & 80.54 & 84.87 & 82.65 & 83.97 & 6590 & 1592 & 8182 \\ \hline
\end{tabular}%
}
\end{table*}

Wang et al.~\cite{wang2020effective}, Lee et al.~\cite{lee2021automated}, and Chen et al.~\cite{chen2021lunar} have the lowest fractional latitude ($4.49$\%), longitude ($5.99$\%), and radial error ($3.99$\%), respectively. 
The fractional errors of all CDAs are approximately the same, ranging from 4.49 to 6.34; this may be because all are using the same labels for training the DL framework, and the same parameters are used in the template matching algorithm. 
Also, we found out that the fractional errors of Silburt et al.~\cite{silburt2019lunar} calculated in our work are significantly less than the errors they mentioned in their paper. It may be due to resizing operation (resizing multiple size DEM to fix size) in their data generation process causes high positional distortion and leads to high fractional errors.

Each CDA architecture has a unique design.  For instance, ERU-Net~\cite{wang2020effective} utilizes residual connections for better training and tackling the vanishing gradient problem. 
HRNet~\cite{chen2021lunar} utilizes multiple upsampling, downsampling, and multi-scale fusion operations to minimize information loss due to downsampling and extract high-level semantic information. CraterUNet~\cite{delatte2019segmentation} has used a larger depth and less number of filters per layer to provide much better space-time complexity.
The feature extraction ability of each CDA framework is different; hence, the crater that one CDA might have missed can be found by another CDA. We combine the detection results of all architectures and eliminate the duplicate craters (Table~\ref{tab:imp_combine}). The recall of this combined detection is $97.71\%$, which is the highest recall of any automated catalog generated using the Povilaitis catalog~\cite{povilaitis2018crater} for training. The unmarked craters in the catalog cause low precision (i.e., $51.73\%$).

\begin{figure*}[htb!]
    \centering
	\includegraphics[width=0.9\linewidth]{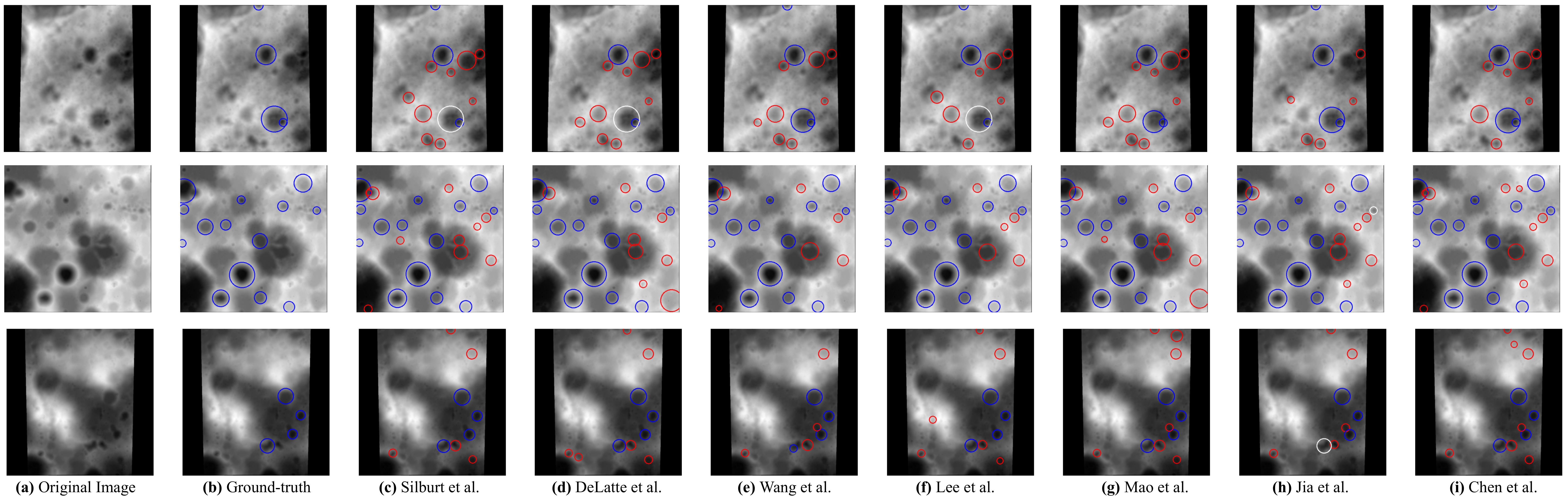}
	\caption{Visual representation of CNN predictions and the corresponding ground truth on lunar DEMs (Blue: detected craters that are present in the ground truth (true positive), Red: detected craters that are not present in the ground truth (false positive), White: undetected craters that are present in the ground truth (false negative)).  \textbf{(a)} Sample DEMs from the test set. \textbf{(b)} Craters present in the ground truth (Povilaitis et al.~\cite{povilaitis2018crater}). \textbf{(c)} Crater predictions generated by DeepMoon~\cite{silburt2019lunar}  which uses the first time U-Net framework~\cite{ronneberger2015u} for crater detection. \textbf{(d)} Detections generated using Crater U-Net~\cite{delatte2019segmentation}, which has a higher depth U-Net with more layers and fewer filters per layer, resulting in a reduced size. \textbf{(e)} Crater predictions generated by ERU-Net network~\cite{wang2020effective} integrated residual connection in U-Net. \textbf{(f)} Crater predictions in Lee et al.~\cite{lee2021automated}, which utilize the RESUNet framework~\cite{zhang2018road}. \textbf{(g)} Crater predictions generated using the Dual-path U-Net framework~\cite{mao2022coupling}. \textbf{(h)} Crater predictions generated using NAUNET network~\cite{jia2021moon} that uses a nested attention gate based U-Net network. \textbf{(i)}  Crater predictions in Chen et al.~\cite{chen2021lunar} obtained using HRNet framework~\cite{sun2019deep} which was originally used for pose estimation.}
	\label{fig:visual_result_sem_seg}
\end{figure*}

The detected craters that are not present in the Povilaitis et al. catalog~\cite{povilaitis2018crater} can be new craters or false positives. The craters detected by more than one CDA will have high certainty to be new craters. Table~\ref{tab:imp_combine} shows that $5202$ craters, which are not in the Povilaitis et al. catalog~\cite{povilaitis2018crater}, are detected by more than one CDA, and $1592$ craters are detected by all CDAs. Hence, these craters can be considered highly certain and can be added to the catalog. 
Also, for the few test images visual results are shown in Figure~\ref{fig:visual_result_sem_seg}.

\paragraph{Overlapping Craters}
The surface of the lunar is covered with craters of varying sizes and shapes, and there can be a scenario in which craters occur in such a way that sufficient separation is not present between two craters; this leads to the formation of overlapping craters~\cite{passey1980effects}.
The study of overlapping craters can give insight into surface erosion and degradation patterns and provide information about which part of the planetary surface is older since there is a higher density of overlapping craters forming on older surfaces, as seen on the highlands on the lunar~\cite{wang2021improved,wilhelms1987geologic}.
One of the challenges in detecting craters is that they degrade over time. With increasing degradation, at some point, the crater may eventually become indistinguishable from the surface. 
The overlapping craters, if sufficiently degraded, may visually resemble a elliptical crater, and this causes a significant problem for crater detection. 
Catalogs such as Robbins et al.~\cite{robbins2019new} and Head et al.~\cite{head2010global} have considered such craters as two separate craters instead of one long elliptical crater. If a crater appears significantly elongated but has a visible cusp in the middle and visually recognizable rims at the ends. In that case, the probability of being correct will be higher if we consider such a feature to be two separate craters instead of one elliptical crater. 

Most traditional methods do not perform well in detecting complex features in craters, such as overlapping craters~\cite{yin2013novel,jia2021split}. However, the deep learning-based method works better than traditional methods. To understand the effectiveness of existing semantic segmentation-based CDAs in overlapping cases, we extracted the overlapping crater from the ground truth (Povilaitis et al.~\cite{povilaitis2018crater}) using the equation provided in Ali-Dib et al.~\cite{ali2020automated} as follows.
\begin{align}
    (r_1 - r_2)^2 < (x_1 - x_2)^2 + (y_1 - y_2)^2 < (r_1 + r_2)^2
\end{align}
Where, ($x_1,y_1$) and ($x_2,y_2$) are the centers and $r_1$ and $r_2$ are the radius of the two craters.

We obtained 808 overlapping craters from ground truth in the test region. 
In Table~\ref{tab:overlap_craters}, we have shown the performance of different CDAs in detecting overlapping craters. It is evident that all methods can detect craters more than 84\% from the ground truth. Silburt et al. and Mao et al. perform best, where Silburt et al. can detect 727, and Mao et al. can detect 728 craters out of 808 overlapping craters in the ground truth.

\begin{table}[!ht]
\centering
\caption{Performance of semantic segmentation-based CDAs in detecting craters out of 808 overlapping craters in the ground truth.}

\label{tab:overlap_craters}
{%
\begin{tabular}{lcc} \hline
\textbf{Existing works} & \textbf{\begin{tabular}[c]{@{}l@{}}Matched Craters\end{tabular}} & \textbf{Recall (\%)}  \\ \hline
Silburt et al.~\cite{silburt2019lunar}        & 727 & 89.97 \\
DeLatte et al.~\cite{delatte2019segmentation} & 693 & 85.77 \\
Wang et al.~\cite{wang2020effective}          & 701 & 86.76 \\
Lee et al.~\cite{lee2021automated}            & 680 & 84.16 \\
Mao et al.~\cite{mao2022coupling}             & 728 & 90.10 \\
Jia et al.~\cite{jia2021moon}                 & 703 & 87.00 \\
Chen et al.~\cite{chen2021lunar}              & 707 & 87.50 \\ \hline
\end{tabular}%
}
\end{table}

\subsection{Object Detection Based CDAs}
\label{sec:Object Detection Based Method}

\begin{figure*}[htb!]
	\includegraphics[width=0.95\linewidth]{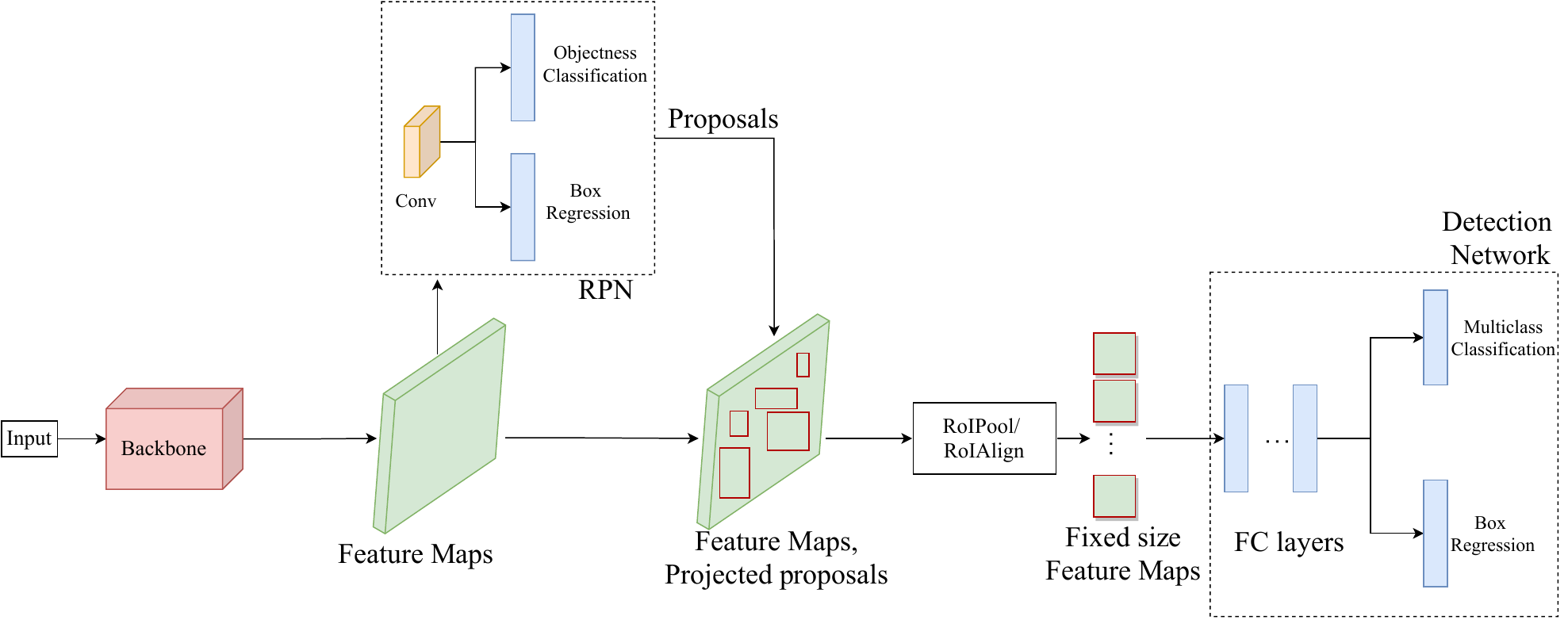}
	\caption{Two Stage Object Detection.}
	\label{fig:two_stage_objd}
\end{figure*}

Object detection aims to detect objects of predefined categories. Most existing object detection-based methods can be divided into one-stage methods~\cite{liu2016ssd,lin2017focal,redmon2017yolo9000} and two-stage methods~\cite{ren2015faster,dai2016r,he2017mask}. 
In a one-stage method, regional proposal extraction and detection will happen simultaneously. In a two-stage method, first region proposals are extracted, then extracted region proposals are used for detection. Two-stage methods typically perform better, whereas single-stage methods have faster inference speed. Most of the existing crater detection methods are based on a two-stage method. One widely used two-stage object detection method is Faster R-CNN~\cite{ren2015faster}. The overview of two-stage object detection is shown in Figure~\ref{fig:two_stage_objd}.
In this, the first features of the image are extracted using convolutional backbone architecture. The extracted features from the backbone are passed to the region proposal network (RPN) to obtain the region proposals, i.e., potential craters' location. This region proposal in the feature map is passed through ROIPool/ROIAlign to get the fixed-sized feature map. Then these proposals are sent to the detection network to get the location and size of the craters. The details of object detection frameworks can be found in the following review works:~\cite{oksuz2020imbalance,liu2020deep,zhao2019object,xiao2020review}.

\subsubsection{Common terminology used in object detection}
\label{subsubsec:Common terminology used in object detection}

\paragraph{Bounding box:}

A rectangular box to define a certain feature in the image (e.g., crater). It is typically represented as ($x_1,y_1,x_2,y_2$), where ($x_1,y_1$) is top-left corner and ($x_2,y_2$) is bottom-right corner.
It is a predefined bounding box used in RPN in two-stage object detection and detection network in one-stage object detection to indicate possible objects of different scales and aspect ratios.

\paragraph{Anchor} 
It is a predefined bounding box that indicates possible objects of different scales and aspect ratios, utilized in RPN in two-stage object detection and detection network in one-stage object detection.

\paragraph{Backbone}
This part is used  to extract the features from the input images.

\paragraph{Region Proposal Network (RPN)}
It is used in two stage object detection networks to extract the region proposals (potential objects) from the extracted features using the backbone. 

\paragraph{Detection Network}
It includes a classifier and regressor to get the bounding box and class of the objects.

\subsubsection{Object detection based method for crater detection}
\label{subsubsec:Object detection based method for crater detection}

Ali-Dib et al.~\cite{ali2020automated} used Mask-RCNN~\cite{he2017mask} to detect craters and extract the shape of the craters. The extracted crater shape was further used to analyze the crater's ellipticity distribution and morphological parameters.

Crater R-CNN with two-teachers self-training with noise (TTSN) was proposed by Zang et al.~\cite{zang2021semi}. The contribution of this work is two-fold. First, inspired by Faster R-CNN~\cite{ren2015faster} and Mask R-CNN~\cite{he2017mask}, Crater R-CNN is proposed. Second, to handle the incompleteness of the data, two-teachers self-training with noise (TTSN) is proposed.
In Crater R-CNN, ResNet with a modified feature pyramid network (FPN) is used in the backbone.
In the FPN~\cite{lin2017feature}, addition is replaced by concatenation in upsampling layers to fuse high semantic
features with high-resolution features.
Similar to Mask R-CNN, ROIAlign is used to generate the fixed-size feature map. 
In TTSN, the incomplete training dataset is first divided into training sets, i.e., $1$ and $2$, and then Gaussian noise is added to both datasets. Crater R-CNN is then trained on training set $1$ to produce teacher model $1$ and on training set $2$ to produce teacher model $2$. Teacher model $1$ is used to perform prediction training set $2$, and teacher model $2$ is used to perform prediction training set $1$. To create a complete dataset, predicted craters with a confidence score greater than $0.75$ are combined with incomplete data.
Finally, the complete training data set was used to train the student model, and performance was assessed using test data.

Hsu et al.~\cite{hsu2021knowledge} integrated the geospatial knowledge into the deep learning process. 
For crater detection, a Faster-RCNN~\cite{ren2015faster} is used as the baseline.
For the feature extractor (backbone), ResNet50~\cite{he2016deep} with FPN~\cite{lin2017feature} is used. Each image and its Hough-transform applied counterpart are parallelly fed into this feature extractor. The output feature maps of the image and Hough transform counterpart are concatenated by $1 \times 1$ convolution.
Then the resultant feature map is then given to the region proposal network, which gives potential crater locations of different scales. This output is sent to the scale-aware classifier, which calculates the probability and weight of the craters. The scale-aware classifier learns two weights to indicate whether the crater is large or small. The weighted sum of category-wise confidence scores is the final score calculated for each crater. This score lies between $0$ and $1$, and if it is $1$, the network is fully confident of the detected crater being an actual crater. 
A threshold of 0.5 is chosen, and if the score
is above, the detection is considered a valid crater.

To address the problem of insufficient feature information from a single data source, Yang et al.~\cite{yang2020lunar} and Jia et al.~\cite{jia2021split} fused the optical image and DEM data for crater detection.
Yang et al.~\cite{yang2020lunar} used R-FCN~\cite{dai2016r} DL framework to detect craters.
Jia et al.~\cite{jia2021split} used a novel split-attention network with self-calibrated convolution (SCNeSt) with FPN to extract features in the R-FCN DL framework. 
Also, Jia et al. replaces position-sensitive ROIPool with position-sensitive ROIAlign. 
SCNeSt modified the ResNest~\cite{zhang2020resnest} by replacing the second convolution layer of the ResNest block with the self-calibration convolution of the SCNet~\cite{liu2020improving} to enhance the diversity of the features. Also, global pooling in the split attention radix group of ResNest is replaced by combining average pooling and max pooling, which obtain better texture and informative features.

To efficiently detect small-size craters, a deep neural network called a high-resolution feature pyramid network (HRFPNet) was proposed by Yang et al.~\cite{yang2021high}. 
In this, first, an adaptive anchor calculation and label assignment (AACLA) algorithm is proposed to collect a sufficient number of anchors for small craters in training; then, these anchors are used in the proposed DL framework, i.e., HRFPNET, to detect the location and size of the craters. 
In AACLA, the first anchors are estimated using the particle swarm optimization algorithm~\cite{marini2015particle}, which collects more training samples of small craters. Then, the label assignment algorithm collects more positive samples of small craters. The HRFPNet consists of a ResNet~\cite{he2016deep} branch and a high-resolution branch. 
ResNet branch is used to obtain global features.
The high-resolution branch can better detect smaller craters since the feature maps resolution does not decrease with an increase in network depth; however, it lacks global features.
To handle this feature aggregation module (FAM) is proposed to enhance the global and contextual features.
Finally, the ResNet branch output is densely connected with the FAM output to get the final feature maps on multiple scales. A total of five feature maps are generated. These are sent to a region proposal network (RPN) and detection network for classification and regression. A focal loss is used for classification, and a balanced regression loss is used for regression. The final predictions from the detection network are filtered using the NMS algorithm to get the size and location of the craters.

Recently, Lin et al.~\cite{lin2022lunar} used a different DL-based object detection framework for crater detection. The following $9$ deep learning frameworks for crater detection are considered: Faster R-CNN~\cite{ren2015faster}, Faster R-CNN with FPN~\cite{lin2017feature}, Cascade R-CNN~\cite{cai2018cascade}, SSD~\cite{liu2016ssd}, RetinaNet~\cite{lin2017focal}, YOLOv3~\cite{redmon2018yolov3}, FoveaBox~\cite{kong2020foveabox}, FCOS~\cite{tian2019fcos}, and RepPoints~\cite{yang2019reppoints}. Overall Faster R-CNN~\cite{ren2015faster} with FPN~\cite{lin2017feature} backbone is performing best for crater detection.

To address the issue of limited labeled datasets, a novel network, CraterDANet, was proposed by Yang et al.~\cite{yang2021craterdanet}. To effectively detect real unannotated data samples in crater detection problems, domain adaptation (DA) is used with auto-annotated synthetic data samples. The source data and labels are synthetic images and corresponding labels. The target data and target labels are real crater images and labels. The source data, source labels, and target data are used to learn a classifier to predict target labels. A lunar crater simulation algorithm~\cite{jian2011fast} is used for source data, i.e., synthetic data generation. The DA has three main components. First, encoder-decoder networks aim to preserve the local structure across domains.
Second, a domain classifier aims to minimize the adversarial loss to learning domain invariant features for target and source domain data. It means that even if ground truth for the source dataset only is provided, the network will learn to detect craters from the domain invariant features of the target dataset since the domain invariant features of both source and target datasets will become similar.
Finally, A multi-scale object detector detects craters from the domain invariant features.

Most research works using object detection CDAs have not published their code, making it challenging to replicate their work in a limited time frame.
Also, object detection-based works are computationally expensive. Consequently, in this work, we have focused on implementing semantic segmentation-based approaches for crater detection. This approach is less computationally expensive and has shown promising outcomes in prior research. Nonetheless, we acknowledge the benefits of object detection-based CDAs and aim to implement them in future work to compare their performance with semantic segmentation-based CDAs.

\subsection{Classification Based CDAs}
\label{sec:Classification Based Method}

In a classification-based method, first, potential crater regions are found using non-deep learning methods such as sliding window and selective search~\cite{uijlings2013selective}. Then these potential crater regions are trained on deep learning-based classification networks to classify crater and non-crater classes. Based on this method, we got a single paper, i.e., Emami et al.~\cite{emami2019crater}. Emami et al.~\cite{emami2019crater} first utilized Hough transform~\cite{yuen1990comparative}, highlight-shadow regions~\cite{urbach2009automatic}, convex grouping~\cite{jacobs1996robust}, and interest points~\cite{323794} algorithms to find the potential craters locations and then utilize convolutional neural networks (CNNs) classification network to classify them into crater and non-crater. Three classification networks are used; first, two have two convolution layers and one fully connected layer, and the last has two convolution layers and three fully connected layers.

The classification-based method heavily relies on selected potential regions of the non-deep learning-based method. Hence, it is less effective than the deep learning-based method~\cite{ren2015faster}. Therefore, most of the recent crater detection works are based on semantic segmentation and object detection-based methods.

\subsection{Evaluation Metrics:}
\label{subsec:Evaluation Metrics}

Evaluation metrics are used for assessing the performance of CDAs. The primary metrics utilized by most of the CDAs are precision (P) and recall (R), which are calculated as follows~\cite{bishop2006pattern,murphy2012machine}:
\begin{align*}
\label{eq:precision_rec}
P &= \dfrac{TP}{TP + FP}\times100\\
R &= \dfrac{TP}{TP + FN}\times100 
\end{align*}
where TP, FP, and FN are the total number of true positives, false positives, and false negatives, respectively. TP is the total number of matches between ground truth and CDA detections, FP is the number of craters detected by the CDA that are not present in the ground truth, and FN is the number of craters missed by the CDA that are present in the ground truth. 

We observed that recall is defined as the ratio of the number of matches between ground truth and CDA detections divided by the total annotated craters in the ground truth. The recall represents the percentage of ground truth craters identified by the CDA. A high recall indicates that there is a low chance that the CDA will miss a true crater (crater present in ground truth). Therefore, recall can be defined as the probability that CDA will identify a true crater.

Similarly, precision is defined as the ratio of the total number of matches between ground truth and CDA detection divided by the total number of craters detected by the CDA. 
The percentage of CDA predictions that are true craters is represented by precision. 
The higher the precision, the less likely the CDA will incorrectly identify a non-crater feature as a crater. Thus, precision can be defined as the probability that the CDA prediction is correct.

We can analyze the CDA's performance based on recall and precision in four scenarios. 
First, low precision and recall imply that CDA misses most true craters, and detections are also incorrect; hence, it has no practical use. 
Second, high precision and high recall imply an ideal CDA that correctly detects all ground-truth craters. 
The third and fourth cases, i.e., high precision and low recall; low recall and high precision, commonly occur in most CDAs.
If precision is low, it means many false craters are detected, and if the recall is low, many true craters are missed. 
Therefore, to give a better idea of the overall performance of a CDA, a harmonic mean of precision and recall, known as the $F_1$-score calculated as:

\begin{align*}
    F_{1}\text{-score} &=  \frac{2\times P\times R}{P + R}
\end{align*}
where P, R denotes precision and recall, respectively. 

When the $F_1$-score is high, it indicates that both precision and recall are high, but when the $F_1$-score is lower, it indicates a greater disparity in precision and recall. 

In application-specific tasks such as automatic rover landing on hazardous surfaces, we prioritize recall over precision because we cannot afford to miss any crater on the lunar surface, which may lead to a rover accident. In such cases, entirely relying on the $F_1$-score is not a good choice. 
Additionally, we can prioritize recall over precision when the catalog used for evaluating CDA is incomplete, i.e., many craters are unmarked in the catalog. That causes low precision if CDA detects new craters. Therefore, to place a greater emphasis on recall, $F_2$-score is calculated as follows:

\begin{align*}
    F_{2}\text{-score} &= \frac{5\times P\times R}{4\times P + R}    
\end{align*}

Finally, in order to assess the network's accuracy in predicting crater size and location, we compute the median fractional errors in latitude, longitude, and radius, which are calculated as follows:
\begin{align*}
    \frac{dLo}{R}&=\frac{|Lo_{P}-Lo_{G}|}{R_GC_{KD}}\\
    \frac{dLa}{R}&=\frac{|La_{P}-La_{G}|}{R_GC_{KD}}\\
    \frac{dR}{R}&=\frac{|R_{P}-R_{G}|}{R_{G}}
\end{align*}

where $Lo$, $La$ and $R$ represent each crater's longitude, latitude, and radius, respectively. 
The subscripts P and G denote predicted and ground-truth measures, respectively.

\subsection{Discussion}
\label{subsec:Discussion}
The working mechanisms of the three CNNs, semantic segmentation, classification, and object detection, are different. They can all be used in suitable pipelines to perform the task of crater detection. In classification-based methods, first potential crater regions are estimated using traditional algorithms such as selective search~\cite{uijlings2013selective}, and then a deep learning-based classification network is used to classify the potential regions into craters and non-craters. It is simpler in structure and easy to implement relative to other categories of crater detection methods, i.e., object detection-based and semantic segmentation-based approaches.

Semantic segmentation-based CDAs segment the image to the crater and non-crater regions. This method provides detailed shape information on the craters. Finally, a template-matching algorithm is used to extract the craters' location and size information. In this approach, images need not be pre-processed to extract the potential crater regions from images like in classification-based crater detection methods. Also, semantic segmentation architectures have a simpler framework that needs to process the image only in one step to get the segmented image compared to object detection-based methods, which need to process one image in multiple steps to extract craters.

Object detection-based CDAs utilize object detection DNNs to give the location and size information of the craters. The object detection-based methods eliminate the need for pre-processing and post-processing steps, which are needed in classification-based and semantic segmentation-based CDAs. The semantic segmentation-based approach provides a precise shape of craters, while the object detection-based approach provides a precise location of the craters. In the future, to utilize the advantage of both, an instance segmentation-based approach can be used to get the precise shape as well as location and size information of the craters. A recent work by Tewari et al.~\cite{tewari2022automatic} has made progress in this direction.

\section{Datasets}
\label{sec:Description of Benchmark Dataset}

\begin{table*}[!ht]
\centering
\caption{Dataset Utilized by Existing Deep Learning-based CDAs.}
\label{tab:dataset_info}
\resizebox{\textwidth}{!}{
\begin{tabular}{|l|l|l|l|ll|ll|l|lll|l|}
\hline
\rowcolor[HTML]{FFFFFF} 
\multicolumn{1}{|c|}{\cellcolor[HTML]{FFFFFF}{\color[HTML]{000000} }} &
  \multicolumn{1}{c|}{\cellcolor[HTML]{FFFFFF}{\color[HTML]{000000} }} &
  \multicolumn{1}{c|}{\cellcolor[HTML]{FFFFFF}{\color[HTML]{000000} }} &
  \multicolumn{1}{c|}{\cellcolor[HTML]{FFFFFF}{\color[HTML]{000000} }} &
  \multicolumn{2}{c|}{\cellcolor[HTML]{FFFFFF}{\color[HTML]{000000} }} &
  \multicolumn{2}{c|}{\cellcolor[HTML]{FFFFFF}{\color[HTML]{000000} }} &
  \cellcolor[HTML]{FFFFFF}{\color[HTML]{000000} } &
  \multicolumn{3}{c|}{\cellcolor[HTML]{FFFFFF}{\color[HTML]{000000} Number of Images}} &
  \cellcolor[HTML]{FFFFFF}{\color[HTML]{000000} } \\ \cline{10-12}
\rowcolor[HTML]{FFFFFF} 
\multicolumn{1}{|c|}{\multirow{-2}{*}{\cellcolor[HTML]{FFFFFF}{\color[HTML]{000000} Paper}}} &
  \multicolumn{1}{c|}{\multirow{-2}{*}{\cellcolor[HTML]{FFFFFF}{\color[HTML]{000000} Region studied}}} &
  \multicolumn{1}{c|}{\multirow{-2}{*}{\cellcolor[HTML]{FFFFFF}{\color[HTML]{000000} Dataset}}} &
  \multicolumn{1}{c|}{\multirow{-2}{*}{\cellcolor[HTML]{FFFFFF}{\color[HTML]{000000} Resolution}}} &
  \multicolumn{2}{c|}{\multirow{-2}{*}{\cellcolor[HTML]{FFFFFF}{\color[HTML]{000000} Catalog}}} &
  \multicolumn{2}{c|}{\multirow{-2}{*}{\cellcolor[HTML]{FFFFFF}{\color[HTML]{000000} \begin{tabular}[c]{@{}c@{}}Craters in \\ Catalog\end{tabular}}}} &
  \multirow{-2}{*}{\cellcolor[HTML]{FFFFFF}{\color[HTML]{000000} \begin{tabular}[c]{@{}l@{}}Image size\\ (pixels)\end{tabular}}} &
  \multicolumn{1}{l|}{\cellcolor[HTML]{FFFFFF}{\color[HTML]{000000} Train}} &
  \multicolumn{1}{l|}{\cellcolor[HTML]{FFFFFF}{\color[HTML]{000000} Validation}} &
  {\color[HTML]{000000} Test} &
  \multirow{-2}{*}{\cellcolor[HTML]{FFFFFF}{\color[HTML]{000000} \begin{tabular}[c]{@{}l@{}}Detected \\ Diameter Range\end{tabular}}} \\ \hline
\rowcolor[HTML]{FFFFFF} 
{\color[HTML]{000000} \begin{tabular}[c]{@{}l@{}}Chen et al.~\cite{chen2021lunar}\\ 2021\end{tabular}} &
  {\color[HTML]{000000} \begin{tabular}[c]{@{}l@{}}Moon\\ Latitude: -60° to 60°\\ Longitude: 0° to 360°\end{tabular}} &
  {\color[HTML]{000000} LRO DEM} &
  {\color[HTML]{000000} 512 pixels/degree} &
  \multicolumn{2}{l|}{\cellcolor[HTML]{FFFFFF}{\color[HTML]{000000} \begin{tabular}[c]{@{}l@{}}Head et al.,  \\ Povilaitis et al.\end{tabular}}} &
  \multicolumn{2}{l|}{\cellcolor[HTML]{FFFFFF}{\color[HTML]{000000} 24,523}} &
  {\color[HTML]{000000} 256 x 256} &
  \multicolumn{1}{l|}{\cellcolor[HTML]{FFFFFF}{\color[HTML]{000000} 30,000}} &
  \multicolumn{1}{l|}{\cellcolor[HTML]{FFFFFF}{\color[HTML]{000000} 5,000}} &
  {\color[HTML]{000000} 5,000} &
  {\color[HTML]{000000} [30, inf) pixels} \\ \hline
\rowcolor[HTML]{FFFFFF} 
{\color[HTML]{000000} \begin{tabular}[c]{@{}l@{}}Jia et al.~\cite{jia2021moon}\\ 2021\end{tabular}} &
  {\color[HTML]{000000} \begin{tabular}[c]{@{}l@{}}Moon\\ Latitude: -60° to 60°\\ Longitude: 0° to 360°\end{tabular}} &
  {\color[HTML]{000000} LRO DEM} &
  {\color[HTML]{000000} 512 pixels/degree} &
  \multicolumn{2}{l|}{\cellcolor[HTML]{FFFFFF}{\color[HTML]{000000} \begin{tabular}[c]{@{}l@{}}Head et al.,\\ Povilaitis et al.\end{tabular}}} &
  \multicolumn{2}{l|}{\cellcolor[HTML]{FFFFFF}{\color[HTML]{000000} 24,523}} &
  {\color[HTML]{000000} 256 x 256} &
  \multicolumn{1}{l|}{\cellcolor[HTML]{FFFFFF}{\color[HTML]{000000} 30,000}} &
  \multicolumn{1}{l|}{\cellcolor[HTML]{FFFFFF}{\color[HTML]{000000} 5,000}} &
  {\color[HTML]{000000} 5,000} &
  {\color[HTML]{000000} [10 , 80] pixels} \\ \hline
\rowcolor[HTML]{FFFFFF} 
{\color[HTML]{000000} \begin{tabular}[c]{@{}l@{}}Wang S. et al.~\cite{wang2020effective}\\ 2020\end{tabular}} &
  {\color[HTML]{000000} \begin{tabular}[c]{@{}l@{}}Moon\\ Latitude: -60° to 60°\\ Longitude: 0° to 360°\end{tabular}} &
  {\color[HTML]{000000} LRO DEM} &
  {\color[HTML]{000000} 512 pixels/degree} &
  \multicolumn{2}{l|}{\cellcolor[HTML]{FFFFFF}{\color[HTML]{000000} \begin{tabular}[c]{@{}l@{}}Head et al.\\ Povilaitis et al.\end{tabular}}} &
  \multicolumn{2}{l|}{\cellcolor[HTML]{FFFFFF}{\color[HTML]{000000} 24,523}} &
  {\color[HTML]{000000} 256 x 256} &
  \multicolumn{1}{l|}{\cellcolor[HTML]{FFFFFF}{\color[HTML]{000000} 30,000}} &
  \multicolumn{1}{l|}{\cellcolor[HTML]{FFFFFF}{\color[HTML]{000000} 3,000}} &
  {\color[HTML]{000000} 3,000} &
  {\color[HTML]{000000} [10 , 80] pixels} \\ \hline
\rowcolor[HTML]{FFFFFF} 
{\color[HTML]{000000} \begin{tabular}[c]{@{}l@{}}Silburt et al.~\cite{silburt2019lunar}\\ 2019\end{tabular}} &
  {\color[HTML]{000000} \begin{tabular}[c]{@{}l@{}}Moon\\ Latitude: -60° to 60°\\ Longitude: 0° to 360°\end{tabular}} &
  {\color[HTML]{000000} LRO DEM} &
  {\color[HTML]{000000} 512 pixels/degree} &
  \multicolumn{2}{l|}{\cellcolor[HTML]{FFFFFF}{\color[HTML]{000000} \begin{tabular}[c]{@{}l@{}}Head et al.,\\ Povilaitis et al.\end{tabular}}} &
  \multicolumn{2}{l|}{\cellcolor[HTML]{FFFFFF}{\color[HTML]{000000} 24,523}} &
  {\color[HTML]{000000} 256 x 256} &
  \multicolumn{1}{l|}{\cellcolor[HTML]{FFFFFF}{\color[HTML]{000000} 30,000}} &
  \multicolumn{1}{l|}{\cellcolor[HTML]{FFFFFF}{\color[HTML]{000000} 30,000}} &
  {\color[HTML]{000000} 30,000} &
  {\color[HTML]{000000} [10 , 80] pixels} \\ \hline
\rowcolor[HTML]{FFFFFF} 
{\color[HTML]{000000} \begin{tabular}[c]{@{}l@{}}Mao et al.~\cite{mao2022coupling} \\ 2022\end{tabular}} &
  {\color[HTML]{000000} \begin{tabular}[c]{@{}l@{}}Moon\\ Latitude: -60° to 60°\\ Longitude: -180° to 180°\end{tabular}} &
  {\color[HTML]{000000} \begin{tabular}[c]{@{}l@{}}LRO DEM\\ LRO WAC\end{tabular}} &
  {\color[HTML]{000000} \begin{tabular}[c]{@{}l@{}}DEM: 512 pixels/degree,\\ Optical: 303.23 pix/deg\end{tabular}} &
  \multicolumn{2}{l|}{\cellcolor[HTML]{FFFFFF}{\color[HTML]{000000} \begin{tabular}[c]{@{}l@{}}Head et al.,\\ Povilaitis et al.\end{tabular}}} &
  \multicolumn{2}{l|}{\cellcolor[HTML]{FFFFFF}{\color[HTML]{000000} 24,523}} &
  {\color[HTML]{000000} 256 x 256} &
  \multicolumn{1}{l|}{\cellcolor[HTML]{FFFFFF}{\color[HTML]{000000} 15,000}} &
  \multicolumn{1}{l|}{\cellcolor[HTML]{FFFFFF}{\color[HTML]{000000} 5,000}} &
  {\color[HTML]{000000} 5,000} &
  {\color[HTML]{000000} [10 , 80] pixels} \\ \hline
\rowcolor[HTML]{FFFFFF} 
{\color[HTML]{000000} \begin{tabular}[c]{@{}l@{}}Lee et al.~\cite{lee2021automated} \\ 2021\end{tabular}} &
  {\color[HTML]{000000} Mars} &
  {\color[HTML]{000000} \begin{tabular}[c]{@{}l@{}}DTM\\ THEMIS Daytime IR\end{tabular}} &
  {\color[HTML]{000000} \begin{tabular}[c]{@{}l@{}}DTM: 200 m/pixel\\ IR: 100 m/pixel\end{tabular}} &
  \multicolumn{2}{l|}{\cellcolor[HTML]{FFFFFF}{\color[HTML]{000000} \begin{tabular}[c]{@{}l@{}}Robbins and \\ Hynek\end{tabular}}} &
  \multicolumn{2}{l|}{\cellcolor[HTML]{FFFFFF}{\color[HTML]{000000} 3,84,343}} &
  {\color[HTML]{000000} 256 x 256} &
  \multicolumn{1}{l|}{\cellcolor[HTML]{FFFFFF}{\color[HTML]{000000} 40,000}} &
  \multicolumn{1}{l|}{\cellcolor[HTML]{FFFFFF}{\color[HTML]{000000} 10,000}} &
  - &
  {\color[HTML]{000000} [10 , 80] pixels} \\ \hline
\rowcolor[HTML]{FFFFFF} 
{\color[HTML]{000000} \begin{tabular}[c]{@{}l@{}}Lee C.~\cite{lee2019automated} \\ 2019\end{tabular}} &
  {\color[HTML]{000000} Mars} &
  {\color[HTML]{000000} DTM} &
  {\color[HTML]{000000} 200 m/pixel} &
  \multicolumn{2}{l|}{\cellcolor[HTML]{FFFFFF}{\color[HTML]{000000} \begin{tabular}[c]{@{}l@{}}Robbins and \\ Hynek\end{tabular}}} &
  \multicolumn{2}{l|}{\cellcolor[HTML]{FFFFFF}{\color[HTML]{000000} 3,84,343}} &
  {\color[HTML]{000000} 256 x 256} &
  \multicolumn{1}{l|}{\cellcolor[HTML]{FFFFFF}{\color[HTML]{000000} 25,000}} &
  \multicolumn{1}{l|}{\cellcolor[HTML]{FFFFFF}{\color[HTML]{000000} 5,000}} &
  {\color[HTML]{000000} 25,000} &
  {\color[HTML]{000000} [10 , 80] pixels} \\ \hline
\rowcolor[HTML]{FFFFFF} 
{\color[HTML]{000000} \begin{tabular}[c]{@{}l@{}}DeLatte et al.~\cite{delatte2019segmentation}\\ 2019\end{tabular}} &
  {\color[HTML]{000000} \begin{tabular}[c]{@{}l@{}}Mars \\ Latitude: -30° to 30°\\ Longitude: 0° to 360°\end{tabular}} &
  {\color[HTML]{000000} THEMIS Daytime IR} &
  {\color[HTML]{000000} 256 pixels/degree} &
  \multicolumn{2}{l|}{\cellcolor[HTML]{FFFFFF}{\color[HTML]{000000} \begin{tabular}[c]{@{}l@{}}Robbins and \\ Hynek\end{tabular}}} &
  \multicolumn{2}{l|}{\cellcolor[HTML]{FFFFFF}{\color[HTML]{000000} 3,84,343}} &
  {\color[HTML]{000000} 512 x 512} &
  \multicolumn{1}{l|}{\cellcolor[HTML]{FFFFFF}{\color[HTML]{000000} -}} &
  \multicolumn{1}{l|}{\cellcolor[HTML]{FFFFFF}{\color[HTML]{000000} -}} &
  {\color[HTML]{000000} -} &
  {\color[HTML]{000000} [4 , 64] km} \\ \hline
\rowcolor[HTML]{FFFFFF} 
{\color[HTML]{000000} \begin{tabular}[c]{@{}l@{}}Jia Y. et al.~\cite{jia2021split}\\ 2021\end{tabular}} &
  {\color[HTML]{000000} \begin{tabular}[c]{@{}l@{}}Moon:\\ Latitude: −65° to 65°\\ Longitude: −180° to 180°\\ \\ Mercury:\\ Latitude: -90° to 90°\\ Longitude: 0° to 360°\\ \\ Mars:\\ Latitude: -90° to 90°\\ Longitude: 0° to 360°\end{tabular}} &
  {\color[HTML]{000000} \begin{tabular}[c]{@{}l@{}}CE-1 DOM,  \\ LRO DEM,\\ Mercury MESSENGER \\ Global DEM ,\\ Mars HRSC and MOLA \\ Blended Global DEM\end{tabular}} &
  {\color[HTML]{000000} \begin{tabular}[c]{@{}l@{}}CE-1 DOM: 120 m/pixel\\ LRO DEM: ~59 m/pixel\\ Mercury DEM: 665 m/pixel \\ Mars DEM: 200 m/pixel\end{tabular}} &
  \multicolumn{2}{l|}{\cellcolor[HTML]{FFFFFF}{\color[HTML]{000000} IAU}} &
  \multicolumn{2}{l|}{\cellcolor[HTML]{FFFFFF}{\color[HTML]{000000} }} &
  {\color[HTML]{000000} \begin{tabular}[c]{@{}l@{}}CE1: \\ 4800 x 4800,\\ 1200 x 1200\\ SLDEM: \\ 1000 x 1000\end{tabular}} &
  \multicolumn{1}{l|}{\cellcolor[HTML]{FFFFFF}{\color[HTML]{000000} 8,000}} &
  \multicolumn{1}{l|}{\cellcolor[HTML]{FFFFFF}{\color[HTML]{000000} 1,000}} &
  {\color[HTML]{000000} 1,000} &
  {\color[HTML]{000000} [0.6 , 860] km} \\ \hline
\rowcolor[HTML]{FFFFFF} 
{\color[HTML]{000000} \begin{tabular}[c]{@{}l@{}}Yang H. et al.~\cite{yang2021craterdanet}\\ 2021\end{tabular}} &
  {\color[HTML]{000000} \begin{tabular}[c]{@{}l@{}}Moon,\\ Latitude: −45° to 46°\\ Longitude: −176.4° to 178.8°\end{tabular}} &
  {\color[HTML]{000000} LRO NAC} &
  {\color[HTML]{000000} 1.5 m/pixel} &
  \multicolumn{2}{l|}{\cellcolor[HTML]{FFFFFF}{\color[HTML]{000000} \begin{tabular}[c]{@{}l@{}}Manually \\ Marked\end{tabular}}} &
  \multicolumn{2}{l|}{\cellcolor[HTML]{FFFFFF}{\color[HTML]{000000} >20,000}} &
  {\color[HTML]{000000} 256 x 256} &
  \multicolumn{2}{l|}{\cellcolor[HTML]{FFFFFF}{\color[HTML]{000000} 12}} &
  {\color[HTML]{000000} 8} &
  {\color[HTML]{000000} [12 , 400] m} \\ \hline
\rowcolor[HTML]{FFFFFF} 
{\color[HTML]{000000} \begin{tabular}[c]{@{}l@{}}Yang S. et al.~\cite{yang2021high}\\ 2021\end{tabular}} &
  {\color[HTML]{000000} \begin{tabular}[c]{@{}l@{}}Mars, Entire surface\\ Moon, Entire surface\end{tabular}} &
  {\color[HTML]{000000} \begin{tabular}[c]{@{}l@{}}MDCD\\ LRNAOC\end{tabular}} &
  {\color[HTML]{000000} \begin{tabular}[c]{@{}l@{}}MDCD: 256 pixels/degree\\ LRNAOC: 1 m/pixel\end{tabular}} &
  \multicolumn{2}{l|}{\cellcolor[HTML]{FFFFFF}{\color[HTML]{000000} \begin{tabular}[c]{@{}l@{}}Manually \\ Marked\end{tabular}}} &
  \multicolumn{2}{l|}{\cellcolor[HTML]{FFFFFF}{\color[HTML]{000000} 12,000}} &
  {\color[HTML]{000000} -} &
  \multicolumn{2}{l|}{\cellcolor[HTML]{FFFFFF}{\color[HTML]{000000} 400}} &
  {\color[HTML]{000000} 100} &
  {\color[HTML]{000000} [6, 250] pixels} \\ \hline
\rowcolor[HTML]{FFFFFF} 
{\color[HTML]{000000} \begin{tabular}[c]{@{}l@{}}Zang et al.~\cite{zang2021semi}\\ 2021\end{tabular}} &
  {\color[HTML]{000000} Moon, Entire surface} &
  {\color[HTML]{000000} CE-2 DOM} &
  {\color[HTML]{000000} 7 m/pixel} &
  \multicolumn{2}{l|}{\cellcolor[HTML]{FFFFFF}{\color[HTML]{000000} \begin{tabular}[c]{@{}l@{}}Manually \\ Marked\end{tabular}}} &
  \multicolumn{2}{l|}{\cellcolor[HTML]{FFFFFF}{\color[HTML]{000000} 41, 614}} &
  {\color[HTML]{000000} 512 x 512} &
  \multicolumn{2}{l|}{\cellcolor[HTML]{FFFFFF}{\color[HTML]{000000} 4,000}} &
  {\color[HTML]{000000} 1,000} &
  {\color[HTML]{000000} < 1 km} \\ \hline
\rowcolor[HTML]{FFFFFF} 
{\color[HTML]{000000} \begin{tabular}[c]{@{}l@{}}Hsu et al.~\cite{hsu2021knowledge}\\ 2021\end{tabular}} &
  {\color[HTML]{000000} Mars, Entire surface} &
  {\color[HTML]{000000} THEMIS Daytime IR} &
  {\color[HTML]{000000} 100 m/pixel} &
  \multicolumn{2}{l|}{\cellcolor[HTML]{FFFFFF}{\color[HTML]{000000} \begin{tabular}[c]{@{}l@{}}Robbins and \\ Hynek\end{tabular}}} &
  \multicolumn{2}{l|}{\cellcolor[HTML]{FFFFFF}{\color[HTML]{000000} 3,84,343}} &
  {\color[HTML]{000000} -} &
  \multicolumn{2}{l|}{\cellcolor[HTML]{FFFFFF}{\color[HTML]{000000} 46,288}} &
  {\color[HTML]{000000} 46,287} &
  {\color[HTML]{000000} [10 , 270] pixels} \\ \hline
\rowcolor[HTML]{FFFFFF} 
{\color[HTML]{000000} \begin{tabular}[c]{@{}l@{}}Yang et al.~\cite{yang2020lunar}\\ 2020\end{tabular}} &
  {\color[HTML]{000000} \begin{tabular}[c]{@{}l@{}}Moon\\ Latitude: −65° to 65°\\ Longitude: 0° to 360°\end{tabular}} &
  {\color[HTML]{000000} \begin{tabular}[c]{@{}l@{}}CE-1 DOM\\ CE-2 DEM\end{tabular}} &
  {\color[HTML]{000000} \begin{tabular}[c]{@{}l@{}}CE-1: 120 m/pixel\\ CE-2: 50 m/pixel\end{tabular}} &
  \multicolumn{2}{l|}{\cellcolor[HTML]{FFFFFF}{\color[HTML]{000000} IAU}} &
  \multicolumn{2}{l|}{\cellcolor[HTML]{FFFFFF}{\color[HTML]{000000} -}} &
  {\color[HTML]{000000} \begin{tabular}[c]{@{}l@{}}CE1: \\ 5000 x 5000,\\ 1000 x 1000\\ CE2: \\ 1000 x 1000\end{tabular}} &
  \multicolumn{1}{l|}{\cellcolor[HTML]{FFFFFF}{\color[HTML]{000000} 5,682}} &
  \multicolumn{1}{l|}{\cellcolor[HTML]{FFFFFF}{\color[HTML]{000000} 1,422}} &
  {\color[HTML]{000000} 791} &
  {\color[HTML]{000000} [0.9, 532) km} \\ \hline
\rowcolor[HTML]{FFFFFF} 
{\color[HTML]{000000} \begin{tabular}[c]{@{}l@{}}Ali-Dib et al.~\cite{ali2020automated}\\ 2020\end{tabular}} &
  {\color[HTML]{000000} \begin{tabular}[c]{@{}l@{}}Moon\\ Latitude: −60° to 60°\\ Longitude: 0° to 360°\end{tabular}} &
  {\color[HTML]{000000} LRO DEM} &
  {\color[HTML]{000000} 59 m/pixel} &
  \multicolumn{2}{l|}{\cellcolor[HTML]{FFFFFF}{\color[HTML]{000000} \begin{tabular}[c]{@{}l@{}}Head et al.,\\ Povilaitis et al.\end{tabular}}} &
  \multicolumn{2}{l|}{\cellcolor[HTML]{FFFFFF}{\color[HTML]{000000} 24,523}} &
  {\color[HTML]{000000} 512 x 512} &
  \multicolumn{2}{l|}{\cellcolor[HTML]{FFFFFF}{\color[HTML]{000000} 1,980}} &
  {\color[HTML]{000000} 70} &
  {\color[HTML]{000000} [5 , 125] km} \\ \hline
\rowcolor[HTML]{FFFFFF} 
{\color[HTML]{000000} \begin{tabular}[c]{@{}l@{}}Lin et al.~\cite{lin2022lunar}\\ 2022\end{tabular}} &
  {\color[HTML]{000000} \begin{tabular}[c]{@{}l@{}}Moon\\ Latitude: −65° to 65°\\ Longitude: 0° to 360°\end{tabular}} &
  {\color[HTML]{000000} LRO DEM} &
  {\color[HTML]{000000} 512 pixels/degree} &
  \multicolumn{2}{l|}{\cellcolor[HTML]{FFFFFF}{\color[HTML]{000000} \begin{tabular}[c]{@{}l@{}}Head et al.,\\ Povilaitis et al.\end{tabular}}} &
  \multicolumn{2}{l|}{\cellcolor[HTML]{FFFFFF}{\color[HTML]{000000} 24,523}} &
  {\color[HTML]{000000} 512 x 512} &
  \multicolumn{3}{l|}{\cellcolor[HTML]{FFFFFF}{\color[HTML]{000000} 17,745}} &
  {\color[HTML]{000000} < 50 km} \\ \hline
\begin{tabular}[c]{@{}l@{}}Emami et al.~\cite{emami2019crater}\\ 2019\end{tabular} &
  Moon, region unspecified &
  LRO NAC &
  1 m/pixel &
  \multicolumn{2}{l|}{\cellcolor[HTML]{FFFFFF}\begin{tabular}[c]{@{}l@{}}Manually \\ Marked\end{tabular}} &
  \multicolumn{2}{l|}{-} &
  600 x 400 &
  \multicolumn{2}{l|}{400} &
  178 &
  [10, 100] m \\ \hline
\end{tabular}%
}
\end{table*}

This section provides an overview of the datasets used for implementing deep learning (DL) based CDAs, along with their comparison in Table~\ref{tab:dataset_info}. 
It provides information such as the region studied,  resolution, and image size used for training DL-based CDAs. 
The most commonly used catalog for lunar surface is the combined catalog of Head et al.~\cite{head2010global} and Povilitis et al.~\cite{povilaitis2018crater}, and for Martian surfaces, Robbins \& Hynek~\cite{robbins2012new} catalog.
The combined catalog of Head et al. and Povilitis et al. contains craters with diameter size $\geq 5$ km, whereas Robbins \& Hynek catalog contains craters with a diameter size  $\geq 1$ km. Some researchers used the IAU catalog to mark the lunar surface craters, including Yang C. et al.~\cite{yang2020lunar} and Jia et al.~\cite{jia2021split}. Unlike others, Yang H. et al.~\cite{yang2021craterdanet}, Yang S. et al.~\cite{yang2021high} and Zang et al. 2021~\cite{zang2021semi} marked the craters manually and used for the training and testing purpose. 
Detailed information of the existing catalogs used in DL based CDAs was provided in Table~\ref{tab:catalog}.
Most CDAs have used image sizes $256\times256$ pixels or $512\times512$ for training the DL framework.
A detailed summary of data utilized for crater detection is provided in Table~\ref{tab:data}; it can be observed that most deep learning methods used DEM data for crater detection.

\subsection{Description of Catalogs used for Training Deep Learning based CDAs}

\begin{table*}[!ht]
\centering
\caption{Catalog Utilize by Existing Deep Learning-based CDAs.}
\label{tab:catalog}
\resizebox{0.8\textwidth}{!}{%
\begin{tabular}{|lllllll|}
\hline
\multicolumn{1}{|l|}{\textbf{Surface}} &
  \multicolumn{1}{l|}{\textbf{Catalog}} &
  \multicolumn{1}{l|}{\textbf{Craters}} &
  \multicolumn{1}{l|}{\textbf{\begin{tabular}[c]{@{}l@{}}Diameter \\ Range (Km)\end{tabular}}} &
  \multicolumn{1}{l|}{\textbf{\begin{tabular}[c]{@{}l@{}}Source \\ Data\end{tabular}}} &
  \multicolumn{1}{l|}{\textbf{\begin{tabular}[c]{@{}l@{}}Resolution\\ (m/pixel)\end{tabular}}} &
  \textbf{\begin{tabular}[c]{@{}l@{}}Manual/Automatic\\ Marking\end{tabular}} \\ \hline
\multicolumn{1}{|l|}{Moon} &
  \multicolumn{1}{l|}{\begin{tabular}[c]{@{}l@{}}IAU$^*$ \\ (1919)\end{tabular}} &
  \multicolumn{1}{l|}{9,137} &
  \multicolumn{1}{l|}{(0, $\infty$)} &
  \multicolumn{1}{l|}{-} &
  \multicolumn{1}{l|}{-} &
  Manual \\ \hline
\multicolumn{1}{|l|}{Moon} &
  \multicolumn{1}{l|}{\begin{tabular}[c]{@{}l@{}}Head et al.~\cite{head2010global}\\ (2010)\end{tabular}} &
  \multicolumn{1}{l|}{5,185} &
  \multicolumn{1}{l|}{[20, $\infty$)} &
  \multicolumn{1}{l|}{LRO DEM} &
  \multicolumn{1}{l|}{474} &
  Manual \\ \hline
\multicolumn{1}{|l|}{Moon} &
  \multicolumn{1}{l|}{\begin{tabular}[c]{@{}l@{}}Povilaitis et al.~\cite{povilaitis2018crater}\\ (2017)\end{tabular}} &
  \multicolumn{1}{l|}{22,746} &
  \multicolumn{1}{l|}{[5, 20]} &
  \multicolumn{1}{l|}{\begin{tabular}[c]{@{}l@{}}LRO DEM, \\ Optical\end{tabular}} &
  \multicolumn{1}{l|}{100} &
  Manual \\ \hline
\multicolumn{1}{|l|}{Moon} &
  \multicolumn{1}{l|}{\begin{tabular}[c]{@{}l@{}}Robbins~\cite{robbins2019new}\\ (2018)\end{tabular}} &
  \multicolumn{1}{r|}{1,296,879} &
  \multicolumn{1}{l|}{[1, $\infty$)} &
  \multicolumn{1}{l|}{\begin{tabular}[c]{@{}l@{}}LRO Optical, \\ DEM, \\ JAXA TC\end{tabular}} &
  \multicolumn{1}{l|}{\begin{tabular}[c]{@{}l@{}}100, \\ 60, \\ 30\end{tabular}} &
  Manual \\ \hline
\multicolumn{1}{|l|}{Mars} &
  \multicolumn{1}{l|}{\begin{tabular}[c]{@{}l@{}}Robbins and Hynek~\cite{robbins2012new}\\ (2012)\end{tabular}} &
  \multicolumn{1}{l|}{3,84,343} &
  \multicolumn{1}{l|}{[1, $\infty$)} &
  \multicolumn{1}{l|}{\begin{tabular}[c]{@{}l@{}}Themis \\ Daytime IR,\\ MOLA Gridded\end{tabular}} &
  \multicolumn{1}{l|}{\begin{tabular}[c]{@{}l@{}}100, 232,\\ 463\end{tabular}} &
  Manual \\ \hline
\multicolumn{1}{|l|}{Mars} &
  \multicolumn{1}{l|}{\begin{tabular}[c]{@{}l@{}}Salamuniccar et al.~\cite{salamuniccar2012lu60645gt}\\ (2012)\end{tabular}} &
  \multicolumn{1}{l|}{1,32,843} &
  \multicolumn{1}{l|}{[2, $\infty$)} &
  \multicolumn{1}{l|}{\begin{tabular}[c]{@{}l@{}}DTM,Thermal \\ infrared IR\end{tabular}} &
  \multicolumn{1}{l|}{-} &
  \begin{tabular}[c]{@{}l@{}}Automatic,\\ Manual\end{tabular} \\ \hline
\multicolumn{7}{l}{$^*$:~\url{http://host.planet4589.org/astro/lunar/}} 
\end{tabular}%
}
\end{table*}

The attempt to produce a complete lunar crater catalog can be traced back to $1982$ when Anderson and Whitaker~\cite{andersson1982nasa} published their lunar catalog of $8,497$ craters. This catalog includes all craters identified by the IAU prior to mid-$1981$, and the remaining craters are all manually marked. This catalog has been approved by NASA and is referred to as the NASA-RP-$1097$ catalog or the AW$82$ catalog.  
An updated version of this catalog containing additional craters identified by IAU is available at~\footnote{\href{http://host.planet4589.org/astro/lunar/}{http://host.planet4589.org/astro/lunar/}} and is commonly referred to as the IAU catalog. This latest update brings the total number of craters in the catalog to $8,639$. It includes the craters' name, longitude, latitude, and diameter value.

The Head et al.~\cite{head2010global} catalog contains $5,185$ manually marked craters with a diameter greater than $20$ km. This catalog attempts to list all visible craters that exhibit a measurable rim and visible central depression. DEMs obtained from the lunar orbiter laser altimeter (LOLA), which was present onboard the lunar reconnaissance orbiter (LRO), were used for marking the craters. The CraterTools extension to ArcMap has been used for measuring the diameter of each crater.

The Povilaitis et al.~\cite{povilaitis2018crater} catalog contains a total of $22,746$ craters with a diameter ranging from $5$ km to $20$ km. The CraterTools extension in ArcGIS was utilized. Craters that have been buried by mares that are barely visible referred to as ghost craters ~\cite{wilhelms1987geologic} are excluded. 
The basemaps used for crater identification are given as follows: first, LROC wide angle camera (WAC) monochrome ($643$  nm) mosaic that has a $60^{\circ}$ average solar incidence and a resolution of $100$ m/pixel~\cite{povilaitis2018crater}, and 
second, a shaded relief map was created by merging LROC WAC digital terrain model (DTM) (GLD100~\cite{scholten2012gld100}) and LOLA polar DTM ($78^{\circ}$ to $90^{\circ}$ N and S)~\cite{smith2010lunar}.

The Robbins catalog~\cite{robbins2019new} contains a total of $20,33,574$ craters, $12,96,879$ of which have diameters greater than $1$ km. This catalog has been marked in two steps; in the first step, the WAC ``morphologic'' mosaic made by the LROC with an average solar inclination of $58^\circ$ was used. The mosaics obtained from WAC at ``dawn'' and ``dusk'' were also used, but because the sun is at the horizon at these times, the smaller craters are obscured by shadows. In the second step, LOLA gridded data record (LOLA GDR), and a merged TC DTM and LOLA mosaic ~\cite{bib:barker2016new} are used to mark craters that were missed due to not being visible in the first step. While marking craters in this catalog, the following assumptions were made: 1) Any lunar feature that appears to have a quasi-circular shape has been marked as a crater; this assumption was made because other geological processes that form circular depressions on the moon are uncommon, and 2) Craters that appeared to be highly elliptical, with a cusp on both sides of the crater rim roughly in the middle of a long axis, were classified as two separate craters and not as one highly elliptical crater.
The craters were marked using ArcMap software, and the basemaps used for crater identification were obtained from the following sources: LRO Camera's (LROC) WAC~\cite{robinson2010lunar}, LRO's LOLA~\cite{smith2010lunar}, and terrain camera (TC) on SELENE~\cite{haruyama2012lunar}.
The liberal marking of craters causes the Robbins catalog to have comparatively higher craters than other lunar catalogs. However, it is argued in Ali-Dib et al.~\cite{ali2020automated} that there is a high possibility for features to be falsely marked as craters.

Robbins and Hynek~\cite{robbins2012new} published a statistically complete catalog of manually identified Martian craters with diameter greater than $1$ km. This catalog lists $3,84,343$ craters, with morphometric information for each crater provided if possible. Two searches were conducted to identify the craters, the first search with THEMIS daytime IR mosaics~\cite{christensen2004thermal} ($100$ m/pixel) and Viking Map ($232$ m/pixel at the equator), and the second search with MOLA images~\cite{smith2001mars,zuber1992mars} ($463$ m/pixel).

The MA132843GT~\cite{salamuniccar2012lu60645gt} catalog is a hybrid catalog (utilizing both manual and automatic crater detection approaches) that has been generated by step-by-step improvements to the Martian catalog MA57633GT~\cite{salamuniccar2008gt}.
MA57633GT is itself a combination of five manually marked catalogs: Barlow et al.~\footnote{\href{http://webgis.wr.usgs.gov/mars.htm}{http://webgis.wr.usgs.gov/mars.htm}}, Rodionova et al.~\footnote{\href{http://selena.sai.msu.ru/Home/Mars_Cat/Mars_Cat.htm}{http://selena.sai.msu.ru/Home/Mars\_Cat/Mars\_Cat.htm}}, Kuzmin~\footnote{\href{http://www.marscraterconsortium.nau.edu/}{http://www.marscraterconsortium.nau.edu/}}, Boyce et al.~\cite{boyce2005ancient},  and~\cite{salamuniccar2006estimation}. 
To generate the MA13283GT catalog from the MA57633GT catalog following steps are taken. First, $57592$ craters are added to this catalog using a traditional CDA on MOLA data resulting in the MA115225GT catalog~\cite{salamuniccar2010method}. 
Second, Salamuniccar et al. 2011b~\cite{salamuniccar2011ma130301gt} extended the MA115225GT catalog using Salamuniccar and Loncaric, 2010a~\cite{salamuniccar2010method} CDA, Bandeira et al.~\cite{bandeira2007impact} CDA, and the MA75919T catalog generated using the Stepinski and Urbach CDA~\cite{stepinski2008completion}.
That resulted in the formation of the MA130310GT catalog. 
Finally, an improved version of Salamuniccar and Loncaric~\cite{salamuniccar2010method} CDA was utilized to get the final MA132843GT catalog.

\begin{table*}[!h]
\centering
\caption{Data utilized by existing deep learning-based CDAs.}
\label{tab:data}
\resizebox{0.6\textwidth}{!}{%
\begin{tabular}{|l|l|l|l|l|l|}
\hline
 &
   &
   &
   &
   &
   \\
\multirow{-2}{*}{\textbf{Surface}} &
  \multirow{-2}{*}{\textbf{Mosaic}} &
  \multirow{-2}{*}{\textbf{Data Type}} &
  \multirow{-2}{*}{\textbf{\begin{tabular}[c]{@{}l@{}}Resolution\\ (meter/pixel)\end{tabular}}} &
  \multirow{-2}{*}{\textbf{Spacecraft}} &
  \multirow{-2}{*}{\textbf{Instrument}} \\ \hline
Moon &
  {\color[HTML]{000000} LRO WAC~\cite{robinson2010lunar}} &
  Optical &
  100 &
  \begin{tabular}[c]{@{}l@{}}Lunar \\ reconnaissance\\ orbiter (LRO)\end{tabular} &
  \begin{tabular}[c]{@{}l@{}}Lunar reconnaissance\\ orbiter camera \\ (LROC)\end{tabular} \\ \hline
Moon &
  {\color[HTML]{000000} LRO LOLA~\cite{smith2010lunar}} &
  DEM &
  59 &
  LRO &
  \begin{tabular}[c]{@{}l@{}}Lunar orbiter\\ laser altimeter\\ (LOLA)\end{tabular} \\ \hline
Moon &
  {\color[HTML]{000000} LRO NAC~\cite{robinson2010lunar}} &
  DEM &
  1.5 &
  LRO &
  \begin{tabular}[c]{@{}l@{}}Narrow angle\\ camera (NAC)\end{tabular} \\ \hline
Moon &
  {\color[HTML]{000000} CE-1 DOM~\cite{zuo2014scientific}} &
  DOM &
  120 &
  CE-1 &
  \begin{tabular}[c]{@{}l@{}}Charge coupled\\ device stereo camera\\ (CCD)\end{tabular} \\ \hline
Moon &
  {\color[HTML]{343434} CE-2 DOM~\cite{zuo2014scientific}} &
  DOM &
  7 &
  CE-2 &
  CCD \\ \hline
Moon &
  {\color[HTML]{000000} CE-2 DEM~\cite{zuo2014scientific}} &
  DEM &
  50 &
  CE-2 &
  CCD \\ \hline
Mars &
  \begin{tabular}[c]{@{}l@{}}Mars MGS \\ MOLA~\cite{fergason2018hrsc}\end{tabular} &
  DTM &
  200 &
  \begin{tabular}[c]{@{}l@{}}Mars global \\ surveyor (MGS),\\ Mars express \\ (MEX)\end{tabular} &
  \begin{tabular}[c]{@{}l@{}}Mars orbiter laser\\ altimeter (MOLA),\\ high-resolution\\ stereo camera (HRSC)\end{tabular} \\ \hline
Mars &
  \begin{tabular}[c]{@{}l@{}}Mars odyssey\\ THEMIS-IR\\ Day~\cite{edwards2011mosaicking}\end{tabular} &
  \begin{tabular}[c]{@{}l@{}}Thermal\\ daytime\\ IR\end{tabular} &
  100 &
  Mars odyssey &
  \begin{tabular}[c]{@{}l@{}}Thermal emission\\ imaging system\\ (THEMIS)\end{tabular} \\ \hline
Mars &
  \begin{tabular}[c]{@{}l@{}}THEMIS-IR \\ Day~\cite{edwards2011mosaicking}\end{tabular} &
  \begin{tabular}[c]{@{}l@{}}Thermal\\ daytime\\ IR\end{tabular} &
  230 &
  Mars odyssey &
  THEMIS \\ \hline
\end{tabular}%
}
\end{table*}

\subsection{Description of Data used for Training Deep Learning based CDAs}

The LRO LOLA DEM mosaic covers the complete longitudinal span of the lunar surface and latitude from 60$^{\circ}$ north to 60$^{\circ}$ south. This DEM was constructed by the LOLA and Kaguya teams from data obtained from the LOLA onboard LRO~\cite{smith2010lunar,bib:barker2016new}. The co-registered data from the SELENE terrain camera was used to correct orbital and pointing geolocation errors. 
This DEM has a resolution of 512 pixels/degree ($\sim$59m/pixel) at the equator with a vertical accuracy between 3 to 4 m. More details of the construction of this DEM mosaic can be found at~\footnote{\href{https://www.usgs.gov/media/images/lro-lola-and-kaguya-terrain-camera-dem-merge-60n60s-512ppd}{https://www.usgs.gov/media/images/lro-lola-and-kaguya-terrain-camera-dem-merge-60n60s-512ppd}}.

The lunar reconnaissance orbiter camera (LROC)~\cite{robinson2010lunar} consisted of three cameras, one wide angle camera (WAC) and two narrow angle cameras (NACs). The WAC contains five filters that capture images in the visible wavelength, i.e., 415, 565, 605, 645, and 690 nm and two filters that capture images in the ultraviolet wavelength, i.e., 320 and 360 nm. 
The WAC captured overlapping images with a resolution of 100 m/pixel.
The NAC captured optical images with a very high resolution of up to 0.5 m/pixel, which are useful for detecting small craters.

The missions Chang'E-1 and Chang'E-2 both had a charged coupled device (CCD) stereo camera and a laser altimeter~\cite{zuo2014scientific}. The CCD camera of Chang'E-1 captured the grayscale images of the lunar surface from $75^{\circ}$ north to $75^{\circ}$ south latitude at a resolution of 120 m/pixel. 
The laser altimeter data of Chang'E-1 of a resolution of 500 m/pixel was used to obtain a DEM model of the lunar surface. 
The DOM images with the highest resolution of 7 m/pixel covering the entire lunar surface have been obtained from the Chang'E-2 CCD camera. 

Themis Day IR Global Mosaic of the Martian surface (Version 2.0) tiles released on November 16, 2006 cover the entire planetary surface with a resolution of 256 pixels per degree ($\sim$230 m/pixel) and can be accessed at~\footnote{\href{https://www.mars.asu.edu/data/thm_dir/}{https://www.mars.asu.edu/data/thm\_dir/}}. Each tile covers $30^{\circ}$ in the latitudinal and longitudinal direction and the total size of all the available images is $92160\times46080$ pixels. An updated version of Themis Daytime Infrared mosaic (version 12) with a resolution of 100 m/pixel~\cite{edwards2011mosaicking} was released in the year 2014 and can be downloaded from~\footnote{\url{ https://astrogeology.usgs.gov/search/map/Mars/Odyssey/THEMIS-IR-Mosaic-ASU/Mars_MO_THEMIS-IR-Day_mosaic_global_100m_v12}}. 

The Martian DEMs captured by altimeters such as MOLA have a fairly low resolution of $463$ m/pixels compared to the lunar DEMs having a resolution of $7$ m/pixel~\cite{zuo2014scientific}. In an attempt to increase the spatial resolution of Martian DEMs, a blended martian DEM mosaic with a resolution of 200 m/pixel was created by combining data obtained from MOLA (463 m/pixel) and  high-resolution stereo camera (HRSC) aboard the Mars express (MEX) spacecraft of the European space agency~\cite{laura2016modeling}~\cite{fergason2017themis}. The HRSC produces multicolor optical images with 10 m/pixel resolution. The three-dimensional nature of images produced by the HRSC makes it possible to derive high-resolution DEMs from the HRSC images with a resolution of around 50 m/pixel. The MOLA DEMs were upsampled, the HRSC DEMs were downsampled to an intermediate resolution of 200 m/pixel, and a blended DEM mosaic was created. The generated mosaic is a hybrid mosaic which is available at~\footnote{\url{https://astrogeology.usgs.gov/search/map/Mars/Topography/HRSC_MOLA_Blend/Mars_HRSC_MOLA_BlendDEM_Global_200mp_v2}}.

\section{Future Direction}
\label{sec:Future Direction}

\subsection{Handling less annotated data}
\label{subsec:Handling less annotated data}
Deep learning algorithms typically perform better with large amounts of data. However, annotating millions of craters is a cumbersome and error-prone task. There is a high degree of disagreement among experts about what constitutes a crater. According to Robbins et al.~\cite{robbins2014variability}, approximately 45\% of experts disagree on what constitutes a crater. Therefore an adequate approach is required to deal with this scenario. One approach is to annotate a small number of highly certain samples of crater samples for training the deep learning algorithm and then increase the number of annotations using a semi-supervised approach. For example, Zang et al.~\cite{zang2021semi} propose a two-teachers self-training with noise (TTSN) method to increase the number of labeled craters in the training dataset. Recently, numerous semi-supervised techniques~\cite{zhou2020learning,chen2020semi,jeong2019consistency,gao2020consistency,bellver2020mask,tang2021proposal} have been developed; these can be used in the future for crater detection.

\subsection{Parameter/Hyper-parameter tuning}
\label{subsec:Parameter/Hyper-parameter tuning}
Most researchers focus primarily on modifying the existing state-of-the-art architectures by introducing additional new layers to increase the depth of the architecture or by adding skip connections to preserve spatial features and other architectural constituents. However, less attention is paid to hyper-parameter tuning. Some examples of such hyper-parameters include learning rate, number of filters in convolutional layers, and kernel size. However, sub-optimal hyper-parameters may result in the model failing to converge and not properly minimizing the loss function, resulting in sub-optimal performance. On the other hand, choosing optimal hyper-parameters can improve performance drastically. For example, in Table~\ref{tab:implementation}, a DeLatte et al.~\cite{delatte2019segmentation} based architecture with no skip or dense connections in the encoder and decoder blocks is comparable to a more complex architecture having residual and/or dense connections in the encoder and decoder blocks. Therefore, it is worthwhile to look into hyper-parameter tuning before proposing significant architectural changes.

\subsection{Accurate shape extraction of the craters}
\label{subsec:Accurate shape extraction of the craters}

Most recent work depicts craters as circular shapes; however, this may not be the exact shape of the craters. 
Figure~\ref{fig:size_shape} shows examples of crater shape variations on the lunar surface. Precisely extracting the shape of the craters can aid in understanding many scientific discoveries. For example, the geometry of craters can be used to visualize the degradation state of craters. If their rims are correctly identified, it is possible to precisely calculate craters' mean diameter, depth, and morphometric characteristics. Additionally, it can help with studies that use morphometric data to classify craters~\cite{kruger2018deriving,chen2017morphological,he2012morphological,zhou2019new,bart2007using}. In previous studies, crater depth and diameter were the most commonly used metrics to define crater morphology and shape. On the other hand, the complex shape of the crater cannot be adequately captured by these two simple metrics measurements. 
Therefore, future research should also focus on developing deep learning based crater shape retrieval methods.

\subsection{Fair comparison}
\label{subsec:Fair comparison}

A typical CDA pipeline consists of the following steps: data generation, pre-processing, deep learning architecture, and post-processing. The specifications selected in the step-by-step process have a  significant impact on the performance of the CDA. When a specific work compares its entire CDA pipeline to the state-of-the-art (SOTA) work, some factors must be considered carefully. For example, the region selected in CDA and SOTA should be the same; then, only we can say with certainty that the performance improvement of CDA is due to the proposed pipeline and not specific to the study region. 
Similarly, we cannot compare one CDA pipeline that detects craters in the 1 to 5 km range to another that detects craters in the 15 to 20 km range. Such a comparison is unfair because the pipeline for detecting craters in the 1 to 5 km range may have been designed primarily for small craters. 
Therefore, the testing region and diameter range chosen in the proposed CDA pipeline and SOTA must be the same.

On the other hand,  if a paper claims that the proposed deep learning architecture performs better than existing works, then the data generation, pre-processing, and post-processing steps involved in the compared CDA pipelines must be the same.  When comparing the effectiveness of deep-learning architectures, it is critical to use the same dataset for training and testing; otherwise, the results may be misleading. For example, if we utilize $512 \times 512$ pixels size images for one architecture and $256 \times 256$ pixels size images for the other. In such cases, the performance variation may be due to a change in the image size and not necessarily due to modification or novelty in the architecture. Therefore, when comparing deep learning architectures in the crater detection process, factors such as the diameter range, training region, number of training images and craters, image resolution, image size feed into the DL architecture, and post-processing parameters must be the same.
From here, we can observe that when we compare deep-learning-based architectures, we need a standard dataset with specified samples for training, validation, and testing. We attempted to do so, and now researchers can validate their proposed deep-learning-based architecture in a lunar surface on a single dataset for comparison if their research focuses on detecting craters ranging in size from 5 to 20 km. However, more research work is needed in this direction.

\subsection{Performance evaluation with respect to particular metrics is misleading}
\label{subsec:Performance evaluation with respect to particular metrics is misleading}

The evaluation of CDA performance using only a single metric may not always be fair. For example, in the case of more conservative catalogs, such as the Head et al.~\cite{head2010global} and Povilitis et al. catalog~\cite{povilaitis2018crater} on the lunar surface, a single metric, such as the $F_1$-score, may be misleading. Many craters are missing in this catalog, so even if the algorithm detects an actual crater, it will be considered a false positive, resulting in low precision and a low $F_1$-score. Similarly, Robbins catalog~\cite{robbins2019new}  has liberally marked the crater on the lunar surface and contains many degraded craters that may be falsely positive; training with such data can confuse the CDA, which is undesirable. Tewari et al.~\cite{tewari2022automated}
attempted to solve the problem of the unmarked crater on the catalog by validating it against another catalog. However, an appropriate procedure is required to evaluate the performance of the different CDA algorithms. Depending on the application, researchers may emphasize recall over precision and vice versa.

\subsection{Difficulty in reproducibility}
\label{subsec:Difficulty in reproducibility}
Incomplete information makes it difficult to replicate the
results of existing SOTA. The researchers should be able to easily replicate a research work, which helps to move the research fast. To avoid the reproducibility problem in CDAs, some important factors, such as detection range in meter and pixel coordinates, spatial resolution, image size fed to the deep learning framework, train-test region, and the number of images and craters used for training and testing need to be provided. Also, in the deep learning framework, information such as learning rate, epochs, optimizer, loss function, and batch size need to be provided. In addition, the code and data must be made publicly available to the research community in order to accelerate the research process and allow researchers to validate their methods on a single data set for a fair comparison.

\section{Conclusion}
\label{sec:Conclusion}

Deep learning (DL) based crater detection methods have gained popularity in recent years due to their ability to learn features on their own and have good generalization capability. 
We have reviewed the DL-based crater detection algorithms (CDAs) and explained the challenges, key characteristics, methods, and datasets used for crater detection. 
We have categorized the DL-based CDAs into three categories: (i) Semantic segmentation,  (ii) Object detection, and (iii) Classification. 
Additionally, Semantic segmentation-based CDAs are trained and tested on a common dataset. Common dataset helps to perform fair comparisons in different DL architectures. 
Finally, we have identified and presented many open issues crucial for crater detection and suggested several promising future directions for developing better crater detection approaches.
In our future works, we plan to explore the implementation of object detection-based CDAs to gain an empirical understanding of their performance.

\section*{Acknowledgment}
This material is based upon work partially supported by the Indian Space Research Organisation (ISRO), Department of Space, Government of India under the Award number ISRO/SSPO/Ch-1/2016-17.
Atal Tewari is supported by TCS Research Scholarship. 
Any opinions, findings, and conclusions or recommendations expressed in this material are those of the author(s) and do not necessarily reflect the views of the funding agencies. Address all correspondence to Nitin Khanna at nitin@iitbhilai.ac.in. 

\bibliographystyle{unsrt}
\bibliography{paper_bib}

\end{document}